\documentclass[10pt,twocolumn,letterpaper]{article}
\usepackage{cvpr}

\definecolor{cvprblue}{rgb}{0.21,0.49,0.74}
\definecolor{cvprgreen}{rgb}{0.2, 0.6, 0.0}
\definecolor{salmon}{RGB}{250,128,114}
\usepackage[pagebackref,breaklinks,colorlinks,allcolors=cvprblue]{hyperref}

\usepackage{algorithm}
\usepackage{algpseudocode}
\usepackage{colortbl, multirow}
\usepackage{xcolor}
\usepackage{amsmath}
\usepackage{pifont}
\usepackage{bm}
\usepackage{pgfplots}
\usepackage{filecontents}
\pgfplotsset{compat=1.18}
\usepackage{breqn}
\usepackage{amssymb}
\usepackage{mdframed}
\usepackage{tcolorbox}

\newcommand{\ourmodel}{DG\textsuperscript{2}CD-Net}

\title{When Domain Generalization meets Generalized Category Discovery:\\ An Adaptive Task-Arithmetic Driven Approach}

\author{Vaibhav Rathore\textsuperscript{1}
\and
Shubhranil B\textsuperscript{1}
\and 
Saikat Dutta\textsuperscript{1,2}\thanks{These authors contributed equally to this work.}
\and 
Sarthak Mehrotra\textsuperscript{1}\footnotemark[1]
\and
Zsolt Kira\textsuperscript{3}
\and
Biplab Banerjee\textsuperscript{1}
\\
\textsuperscript{1}IIT Bombay \quad \textsuperscript{2}IITB-Monash Research Academy \quad \textsuperscript{3}Georgia Institute of Technology
}

\begin{document}
\maketitle

\begin{abstract}
Generalized Class Discovery (GCD) clusters base and novel classes in a target domain, using supervision from a source domain with only base classes. Current methods often falter with distribution shifts and typically require access to target data during training, which can sometimes be impractical. To address this issue, we introduce the novel paradigm of Domain Generalization in GCD (DG-GCD), where only source data is available for training, while the target domain—with a distinct data distribution—remains unseen until inference. 
To this end, our solution, \ourmodel, aims to construct a domain-independent, discriminative embedding space for GCD. The core innovation is an episodic training strategy that enhances cross-domain generalization by adapting a base model on tasks derived from source and synthetic domains generated by a foundation model. Each episode focuses on a cross-domain GCD task, diversifying task setups over episodes and combining open-set domain adaptation with a novel margin loss and representation learning for optimizing the feature space progressively. 
To capture the effects of fine-tunings on the base model, we extend task arithmetic by adaptively weighting the local task vectors concerning the fine-tuned models based on their GCD performance on a validation distribution. This episodic update mechanism boosts the adaptability of the base model to unseen targets. Experiments across three datasets confirm that \ourmodel \ outperforms existing GCD methods customized for DG-GCD.
\end{abstract}

\vspace{-10px}

\section{Introduction}
\label{sec:intro}

Recent advances in deep learning \cite{voulodimos2018deep, chai2021deep} have improved visual inference models; however, challenges persist in limited supervision, domain shifts, and detecting novel classes in inference. Semi-supervised and self-supervised methods \cite{sem-survey, ssl-survey, sslssl} address limited supervision but assume a closed-set environment. Domain adaptation (DA) and domain generalization (DG) \cite{da-survey, dg-survey1} aim to create domain-agnostic embeddings for distribution generalization. However, in open-set applications \cite{saito2018open, ODG3}, they typically classify novel samples in a single \texttt{outlier} class.

\begin{figure}[!t]
    \centering
    \includegraphics[width=\columnwidth]{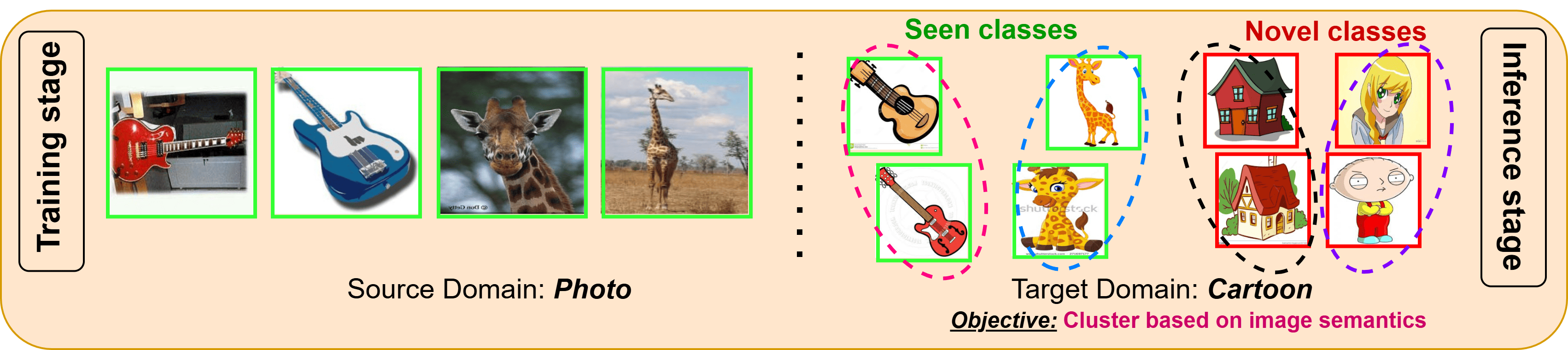}
    \caption{We present a novel variant of GCD, \textbf{Domain Generalization for Generalized Category Discovery}, where a model is trained on a \textbf{source domain} (photo) and evaluated on a \textbf{target domain} (cartoon). During inference, the model must cluster \textbf{seen classes} while also identifying and distinguishing \textbf{novel classes}.}
    \label{fig:dgcd_setting}
    \vspace{-18px}
\end{figure}

Applications like autonomous driving and healthcare require finer categorization of novel samples. Novel Class Discovery (NCD) \cite{hsu2017learning, ncd} addresses this by clustering novel classes in an unlabeled target domain using labeled source data with known classes. Generalized Class Discovery (GCD) \cite{gcd, gcd1, gcd2} extends this to both known and novel classes in the target domain, with the support of source domain knowledge. Cross-Domain GCD (CD-GCD) \cite{rongali2024cdadnetbridgingdomaingaps} further advances this by handling cases where the source and target domains differ in distribution. Nonetheless, all these settings assume simultaneous access to source and target data to carry out training, which is often impractical. For example, an autonomous vehicle \cite{sun2022shift,zhang2019unseen} trained on California’s roads in summer may face unexpected challenges on snowy Moscow streets.
Similarly, medical AI models \cite{yoon2023domain,behar2023generalization} trained on high-quality images may need to analyze lower-resolution scans from rural areas to detect new pathologies. In these cases, the target domains are majorly inaccessible until inference. Thus, robust AI systems must adapt to varying environments, organizing both known and novel data categories without prior exposure and ensuring reliable performance in unexpected scenarios.


To meet these challenges, we propose DG-GCD (Table \ref{fig:dgcd_setting}), a novel and practical setting that leverages a labeled source domain to train a feature extractor on known classes. The task is to leverage the trained model to cluster unlabeled samples from a previously unseen, visually distinct target domain containing both known and novel classes, with an undefined known-novel class ratio—suggesting \textit{solving the GCD problem within a DG framework with uncharted domain and semantic shifts between training and testing}.

Existing GCD and CD-GCD methods perform sub-optimally in DG-GCD, when we train them solely on the source domain without considering the target domain as per DG requirements (Table \ref{tab:results}). Besides, single-source open-set DG methods \cite{ODG3}, which generalize from a labeled source to a mixed-class target domain, categorize all novel samples as a single outlier, diverging from DG-GCD’s objectives.

\begin{figure}[!t]
    \centering
    \includegraphics[width=0.75\linewidth]{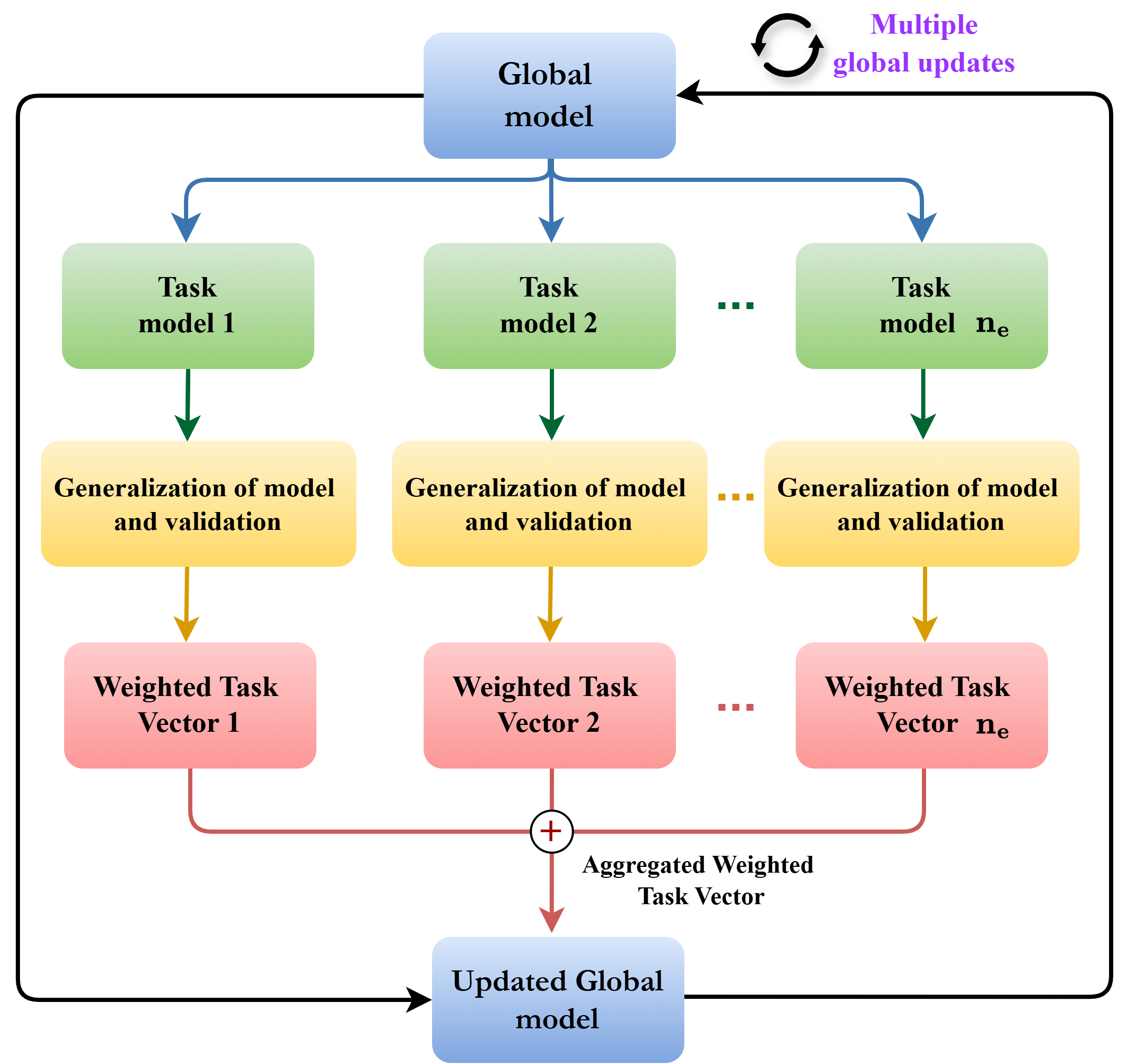}
    \caption{\textbf{Proposed episodic training}: A pre-trained global model is updated using task vectors from $n_e$ episode-specific fine-tuned models, leveraging a novel dynamic weighting scheme. This scheme adjusts the task vectors based on their GCD generalization performance on a held-out unseen validation distribution.}
    \label{fig:teaser1}
    \vspace{-10px}
\end{figure}


\noindent \textbf{Our proposal:} As a remedy, we present DG-GCD-Net (\textit{aka} {\ourmodel}). Generalizing from a single-source domain is inherently challenging due to limited data diversity, the need to adapt to unseen novel classes in the target domain, and the varying known-novel class splits that complicate learning. Standard approaches often struggle to effectively handle both domain shifts and class-level variations.

To address this, we propose a novel \textbf{progressive episodic training strategy} (Fig. \ref{fig:teaser1}) designed to enhance a pre-trained global feature encoder. In each episode, the CD-GCD task is tackled locally by using a subset of the source domain with a reduced class set. Additionally, a stylistic variant of the source domain, generated via a diffusion model \cite{brooks2023instructpix2pix}, acts as a pseudo-target domain containing all known classes, but is treated unlabeled. These episodic splits are dynamically generated to simulate diverse conditions.

\definecolor{ArtBar}{RGB}{142,124,195}
\definecolor{ClipartBar}{RGB}{166,77,121}
\definecolor{ProductBar}{RGB}{69,129,142}
\definecolor{AugBar}{RGB}{246,178,107}
\definecolor{SynBar}{RGB}{106,168,79}

\definecolor{Ties}{RGB}{128,128,192}
\definecolor{Fisher}{RGB}{192,128,128}
\definecolor{TaskArth}{RGB}{128,192,128}
\definecolor{ours}{RGB}{192,192,128}

\begin{figure}[!t]
    \centering
    \resizebox{0.90\linewidth}{!}{
    \begin{tikzpicture}
        \begin{axis}[
            ybar,
            width=\linewidth,
            height=0.5\linewidth,
            bar width=18pt,
            xlabel={},
            xlabel style={font=\footnotesize},
            symbolic x coords={Ties Merging, Fisher Merging, Task Arithmetic, DG$^{2}$CD-Net},
            xtick=\empty,
            xticklabel style={font=\scriptsize, anchor=east},
            ytick={0.25,0.50,0.75,1.00}, 
            yticklabel style={font=\scriptsize},
            ylabel={Fraction of sign conflicts},
            ylabel style={font=\footnotesize},
            grid=none,
            ymin=0,
            ymax=1.2, 
            ymajorgrids=true,
            nodes near coords,
            nodes near coords style={
                font=\scriptsize, 
                inner ysep=2pt 
            },
            enlarge x limits=1.15, 
            axis line style={thick}, 
            grid style={line width=.1pt, draw=gray!10}, 
            major grid style={line width=.2pt,draw=gray!50},
            legend style={
                at={(0.5,1.03)}, 
                anchor=south, 
                legend columns=-1,
                font=\scriptsize, 
            },
            legend image code/.code={
                \draw[#1, draw=none] (0.0cm,-0.1cm) rectangle (0.15cm,0.15cm);
            },
            ]
            \addplot[fill=SynBar, draw=none] coordinates {(Ties Merging,0.9645140719)};
            \addplot[fill=ProductBar, draw=none] coordinates {(Fisher Merging,0.6283945308)};
            \addplot[fill=ClipartBar, draw=none] coordinates {(Task Arithmetic,0.5837368395)};
            \addplot[fill=ArtBar, draw=none] coordinates {(DG$^{2}$CD-Net,0.5442273788)};
            
            \legend{Ties Merging \cite{yadav2024ties}, Fisher Merging \cite{matena2022merging}, Task Arithmetic \cite{task_arithmetic}, DG$^{2}$CD-Net}
        \end{axis}
    \end{tikzpicture}}
    \vspace{-2pt}
    \caption{\textbf{Sign Conflicts}—opposing gradient updates leading to greater parameter divergence across episodes in various \textbf{model merging methods} for our episodic training on PACS. Lower values indicate improved training stability and accuracy.}
    \label{fig:signconflicts}
    \vspace{-15px}
\end{figure}
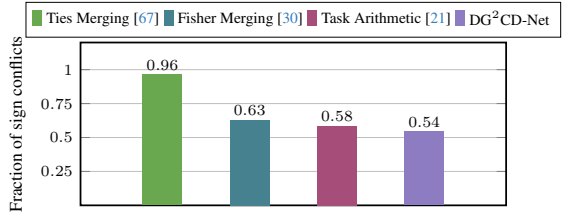

For CD-GCD, known classes from the source (labeled) and synthesized (unlabeled) domains are aligned using an open-set domain adaptation (OSDA) objective, while contrastive learning boosts feature discriminativeness. Existing OSDA methods like \cite{saito2018open} often fail with fine-grained novel classes, causing misalignment with known classes. To address this, we propose a \textbf{novel margin objective} to enforce clearer separation between episode-specific known and novel classes, improving novel class discovery.

To aggregate task-specific knowledge from multiple fine-tuned models, we apply task arithmetic (TA) \cite{task_arithmetic}, combining task vectors that capture weight shifts between pre-trained and fine-tuned models to improve adaptability. While certain TA variants focus only on top-ranked parameters \cite{yadav2024ties}, this approach leads to instability in our context of multiple parallel episodes with varied data distributions. To maintain consistency, we use TA across the entire weight space. 
To enhance generalization beyond traditional TA, we introduce a novel \textbf{dynamically weighted TA} approach tailed for DG-GCD, which deals with scaling task vectors according to each fine-tuned model's generalization performance, assessed through clustering on a diverse validation set. This prioritizes broad generalization, preventing over-specialization and supporting robust adaptation across novel domains. In contrast to approaches like \cite{matena2022merging}, which rely on the Fisher matrix for weighted model merging, our approach better accommodates divergent data distributions, making it more suited to our training scheme (Fig. \ref{fig:signconflicts}, Table \ref{tab:ablation}).
 Our major contributions are summarized as follows:

\noindent \textbf{[-]} We introduce a novel variant of GCD, DG-GCD, where the target domain, distinct from the source in distribution, is only encountered at evaluation.

\noindent \textbf{[-]} We introduce an innovative episodic training strategy that utilizes adaptive task vector modeling, dynamically updating the pre-trained global model by leveraging the generalizability of fine-tuned models tailored for CD-GCD.

\noindent \textbf{[-]} We perform extensive experiments on three datasets, benchmarking our model against several baselines and conducting thorough ablation studies.

\section{Related Works}
\paragraph{Domain generalization:} DG \cite{dg-survey1, dg-survey2} trains models on one or more source domains to generalize to unseen target domains, covering scenarios like closed-set, open-set, multi-source, and single-source. Multi-source DG enhances model robustness using techniques like invariant feature learning and meta-learning \cite{closedDG1, closedDG3, closedDG4, closedDG2, closedDG5, closedDG6}, while single-source DG \cite{SDG1, SDG2, SDG3, SDG4} relies on data augmentation and regularization due to limited domain diversity.

Closed-set DG assumes all target domain classes are seen during training, whereas open-set DG \cite{ODG1, ODG2, ODG3, ODG4,bose2023beyond,wang2023generalizable,peng2025advancing} introduces unseen target classes, increasing complexity. Recent methods, such as \cite{odgclip}, use prompt learning with CLIP to address open-set DG. However, open-set DG groups all novel samples into a single class, unlike our DG-GCD framework, which aims for fine-grained discovery and clustering of novel classes.

\noindent \textbf{(Cross-domain) class discovery:} While both NCD and GCD are well-studied problems \cite{hsu2017learning, ncd1, ncd2, ncd3, ncd4, gcd1, gcd2, gcd3, gcd4} (see Section \ref{sec:lit_review} for more details), class discovery under domain shift remains relatively new. Emerging methods such as CDAD-Net \cite{rongali2024cdadnetbridgingdomaingaps}, Exclusive Style Removal \cite{wang2024exclusivestyleremovalcross}, CROW \cite{wen2024cross}, and HiLo \cite{wang2024hilolearningframeworkgeneralized} concentrate on aligning known classes across domains and strengthening the generalization of novel class discovery amid distributional shifts. These approaches, assuming target domain presence during training, serve as upper bounds for {\ourmodel}'s performance.

\noindent \textbf{Model aggregation:} Fine-tuned models initialized from the same pre-trained model often follow similar optimization paths, making them suitable for merging \cite{ma1, ma2, ma3, task_arithmetic}. Model merging has been applied across various fields, including out-of-domain generalization \cite{ma3, task_arithmetic, swad, daensemble}, multitask learning \cite{ma3}, federated learning \cite{li2019convergence}, multimodal merging \cite{sung2023empirical}, and continual learning \cite{yadav2022exclusive, yadav2023exploring}.

Task Arithmetic \cite{task_arithmetic} improves merging by using task vectors—differences between the weights of a pre-trained model and its fine-tuned versions—to update the base model. By integrating these task vectors, TA aims to boost the pre-trained model’s performance on new tasks while preserving stability. Further refinements to TA have been proposed, such as Ortiz-Jimenez \etal \cite{ortiz2024task}, using a weight disentanglement framework in tangent space, and Yadav \etal \cite{yadav2024ties}, addressing parameter interference during merging.
{\ourmodel} \ extends on the conventional TA by introducing an episodic training scheme to handle dynamic domain and semantic shifts, which static updates in \cite{task_arithmetic, yadav2024ties} fail to address. We also adaptively weigh task vectors based on the GCD generalizability of the fine-tuned models on held-out and disjoint data distributions, ensuring the global model avoids bias towards subpar local models.

\vspace{-5px}
\section{Problem Definition \& Methodology}

In this section, we introduce the DG-GCD problem, where a model trained solely on labeled data from a source domain \(\mathcal{S}\) must identify and cluster both known and novel categories in an unlabeled target domain \(\mathcal{T}\) during testing, without prior exposure, despite domain shifts (\(\mathcal{P}(\mathcal{S}) \neq \mathcal{P}(\mathcal{T})\)).
The source domain dataset is \(\mathcal{D}_{\mathcal{S}} = \{(x_i^s, y_i^s)\}_{i=1}^{n_s}\), where \(x_i^s \in \mathcal{X}_s\) are inputs and \(y_i^s \in \mathcal{Y}_s\) are labels of known categories. The target domain dataset \(\mathcal{D}_{\mathcal{T}} = \{x_j^t\}_{j=1}^{n_t}\) contains both known and novel categories, denoted as $\mathcal{Y}_{t}^{old} = \mathcal{Y}_s$ and $\mathcal{Y}_{t}^{new} = \mathcal{Y}_{t} \setminus \mathcal{Y}_s$, respectively.
The core challenge is to develop a domain-generalizable and discriminative embedding space suitable for on-the-fly GCD evaluation.


\subsection{The \textsc{{\ourmodel}} model}

We introduce {\ourmodel}, initialized from a pre-trained encoder architecture $\mathcal{F}^0$ with parameters $\theta_{\text{global}}$ based on the Vision Transformer (ViT) \cite{vit}. To enhance the generalization capabilities of $\theta_{\text{global}}$ for GCD tasks, we propose an adaptive task-vector aggregation strategy within an episodic training framework. This framework leverages both source data $\mathcal{D}_{\mathcal{S}}$ and a set of $\mathcal{N}$ synthetic pseudo-domains \(\mathcal{D}_{\text{syn}}\), generated by a pre-trained foundation model to systematically vary style elements while preserving the semantic structure of $\mathcal{D}_{\mathcal{S}}$. Our training scheme with such varied data aims to progressively strengthen the domain independence and discriminative capacity of $\theta_{\text{global}}$. Further details on the domain synthesis strategy are provided in Section \ref{sec:syn}.

Our training protocol comprises multiple rounds of global updates for $\theta_{\text{global}}$, each consisting of several local episodes. In each episode \(e\) within a global update \(g\), we select a labeled subset \(\mathcal{D}_{\mathcal{S}}^{e_g} \subset \mathcal{D}_{\mathcal{S}}\) and a randomly chosen synthetic domain \(\mathcal{D}_{\text{syn}}^{e_g}\), enabling the model to adapt locally to the CD-GCD task \cite{rongali2024cdadnetbridgingdomaingaps} and thereby account for distribution shifts in a GCD context. In this setup, the subset \(\mathcal{D}_{\mathcal{S}}^{e_g}\) contains a reduced set of classes, \(\mathcal{Y}_{s}^{e_g} \subset \mathcal{Y}_{s}\) and is treated labeled, while \(\mathcal{D}_{\text{syn}}^{e_g}\) includes the full class set $\mathcal{Y}_{s}$ but is assumed unlabeled. $\mathcal{Y}_{s} \setminus \mathcal{Y}_{s}^{e_g}$ defines the episode-specific novel classes in $\mathcal{D}_{\text{syn}}^{e_g}$. Varying these subsets over episodes exposes the model to diverse domain and semantic shifts, progressively aiding in enhancing the robustness of the updated $\theta_{\text{global}}$ for DG-GCD.

To address the CD-GCD problem, we apply task-specific objectives to each episode pair \((\mathcal{D}_{\mathcal{S}}^{e_g}, \mathcal{D}_{\text{syn}}^{e_g})\) initialized from the current global model, $\theta_{\text{global}}^{g-1}$, yielding an episode-specific parameterization \(\theta_{\text{local}}^{e_g}\). The generalization performance of \(\theta_{\text{local}}^{e_g}\) is evaluated on a validation set, \({\mathcal{D}}_{\text{valid}}\), constructed from diverse synthetic domains distinct from those used in the episodes and held constant throughout training. Keeping \({\mathcal{D}}_{\text{valid}}\) a mixed distribution and separate helps assess the DG-GCD ability for the local models more effectively in a challenging scenario (see Table \ref{tab:ablation} for ablation).
We assess performance on \({\mathcal{D}}_{\text{valid}}\) using K-means clustering with Hungarian assignment \cite{kuhn1955hungarian}. The resulting validation metrics guide the adaptive weighting of task vectors for \(\theta_{\text{local}}^{e_g}\), and the knowledge from all episodes within the global update $g$ is aggregated to refine $\theta_{\text{global}}^{g-1}$ into $\theta_{\text{global}}^{g}$ (or $\mathcal{F}^g$). The complete training process, together with the CD-GCD losses, are detailed in Section \ref{sec:train}.

\subsubsection{Synthetic domain generation }\label{sec:syn}

Our aim is to generate diverse synthetic domains, $\mathcal{D}_{\text{syn}}$, to introduce distributional variety that enhances generalization beyond the source domain, $\mathcal{D}_{\mathcal{S}}$. Manual ad-hoc augmentations, however, are limited by the need for domain-specific expertise and lack the flexibility to introduce realistic distribution shifts. To address this, we leverage Instruct-Pix2Pix \cite{brooks2023instructpix2pix}, a pre-trained conditioned diffusion model, which enables controlled and varied transformations via targeted prompts. This approach allows for modifications in texture, lighting, and environmental factors while preserving semantic integrity, resulting in realistic and adaptable distribution shifts (see Table \ref{tab:ablation} for comparisons). 
Moving forward, we start with curating a number of domain descriptors that can introduce such variations in $\mathcal{D}_{\mathcal{S}}$ by querying ChatGPT-4o \cite{openai2024chatgpt} as follows:

\begin{center}
\begin{tcolorbox}[
    width=0.85\linewidth, 
    colback=white!95!gray,       
    colframe=blue!60!black,      
    coltitle=blue!40!black,      
    boxrule=0.5pt,               
    fonttitle=\bfseries,         
    arc=5pt,                     
    sharp corners=north          
]
{\textit{Generate a list of prompts in the format ``Add [description] scene to the image." Replace `[description]' with single-word terms or short phrases representing colors, weather, time of day, or location. Ensure variety with minimal adjectives. Each should evoke a distinct visual trait.}}
\end{tcolorbox}
\end{center}

\begin{figure}[t!]
    \centering
    \begin{tikzpicture}
        \begin{axis}[
            width=0.98\linewidth, 
            height=4.5cm,
            ybar=0pt, 
            symbolic x coords={Art, Clipart, Product, Avg ad-hoc Augmentations, Avg Synthetic},
            xlabel={Source: \textbf{Real World}},
            ylabel={FID Score},
            ymin=0,
            ymax=125,
            bar width=18pt, 
            nodes near coords,
            nodes near coords style={
                font=\scriptsize, 
                inner ysep=2pt 
            },
            enlarge x limits=1.15, 
            axis line style={thick}, 
            grid style={line width=.1pt, draw=gray!10}, 
            major grid style={line width=.2pt,draw=gray!50}, 
            grid=none, 
            ymajorgrids=true, 
            xticklabels={}, 
            xtick=\empty,
            ytick={0,25,50,75,100}, 
            yticklabel style={font=\scriptsize},
            legend style={
                at={(0.5,1.03)}, 
                anchor=south, 
                legend columns=-1,
                font=\scriptsize, 
            },
            legend image code/.code={
                \draw[#1, draw=none] (0.0cm,-0.1cm) rectangle (0.15cm,0.15cm);
            }
            ]
            \addplot[fill=ArtBar, draw opacity=0] coordinates {(Art, 39.73)};
            \addplot[fill=ClipartBar, draw opacity=0] coordinates {(Clipart, 54.33)};
            \addplot[fill=ProductBar, draw opacity=0] coordinates {(Product, 30.65)};
            \addplot[fill=AugBar, draw opacity=0] coordinates {(Avg ad-hoc Augmentations, 11.955)};
            \addplot[fill=SynBar, draw opacity=0] coordinates {(Avg Synthetic, 107.8)};
            \legend{Art, Clipart, Product, Avg ad-hoc Augmentations, Avg Synthetic}
        \end{axis}
    \end{tikzpicture}
    \vspace{-10pt}
    \caption{The \textbf{FID scores} \cite{eiter1994computing} between the \textit{Real-World} and the three other Office-Home \cite{officehome} domains, along with the average FID between \textit{Real-World} and the InstructPix2Pix-driven and manually-crafted synthesized domains, confirm that the InstructPix2Pix-driven synthetic domains introduce diverse distribution shifts into our episodic training process, leading to enhanced generalizability of $\theta_{\text{global}}$ through the proposed training scheme.}
    \label{fig:main_diag}
    \vspace{-10px}
\end{figure}
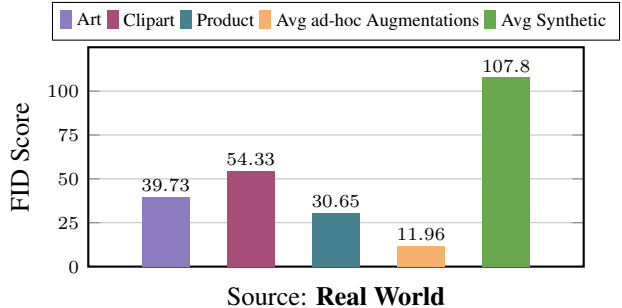

This generates domain primitives such as \texttt{Snowy-Scene} and \texttt{Rainy-Weather} for textures and occlusions, \texttt{Night-Time} for low visibility, and \texttt{Beach}, \texttt{Forest}, and \texttt{Urban} for environmental diversity. These curated prompts are further used in Instruct-Pix2Pix to synthesize the images. Importantly, this approach avoids leaking any target-domain-specific knowledge, preserving DG integrity. As shown in Fig.~\ref{fig:main_diag}, these synthesized domains enhance the distribution diversity of $\mathcal{D}_{\mathcal{S}} \cup \mathcal{D}_{\text{syn}}$, more effectively surrogating the original target counterparts than manual ad-hoc image augmentations (see Sections \ref{sec:synthetic_data} and \ref{sec:synthetic_utilization} for further quantitative and qualitative analyses).

\begin{figure*}[!t]
  \centering
  \includegraphics[width=0.85\linewidth]{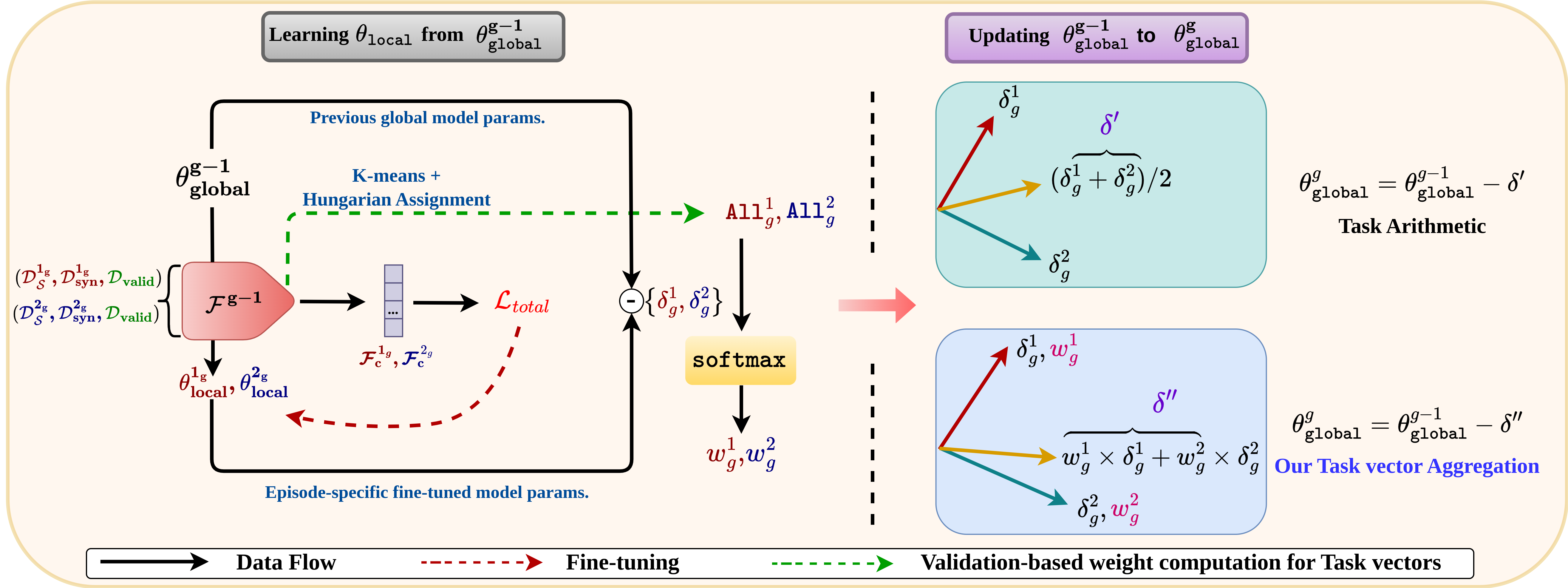}
  \caption{\textbf{Transition from $\theta_{\text{global}}^{g-1}$ to $\theta_{\text{global}}^{g}$ in our training strategy:} The \textbf{left panel} illustrates the two-way episodic training process. Starting with episode-specific datasets $(\mathcal{D}_{\mathcal{S}}^{1_g}, \mathcal{D}^{1_g}_{\text{syn}})$ and $(\mathcal{D}_{\mathcal{S}}^{2_g}, \mathcal{D}^{2_g}_{\text{syn}})$, we fine-tune the previous global model $\mathcal{F}^{g-1}$ together with episode-specific adversarial classifiers $\mathcal{F}_c^{1_g}$ and $\mathcal{F}_c^{2_g}$ on the local CD-GCD tasks. This produces fine-tuned models with updated weights $\theta_{\text{local}}^{1_g}$ and $\theta_{\text{local}}^{2_g}$. We calculate the task vectors ($\delta_g^1, \delta_g^2$) for the fine-tuned models. GCD generalization is subsequently assessed on $\mathcal{D}_{\text{valid}}$ using the \texttt{All} metric, resulting in generalization weights $(w_g^1, w_g^2)$ for the fine-tuned models. The \textbf{right panel} shows how the global models are updated through task vector aggregations for baseline TA \cite{task_arithmetic} and ours. \textbf{Red} and \textbf{Blue} denote the episodes-specific data/processing.}   
  \label{fig:model1}
  \vspace{-15px}
\end{figure*}

\subsubsection{Proposed episodic training strategy using task vectors to update $\theta_{\text{global}}$ and the CD-GCD objectives} \label{sec:train}
Given datasets $\mathcal{D}_{\mathcal{S}}$ and $\mathcal{D}_{\text{syn}}$, our goal is to optimize $\theta_{\text{global}}$ to construct a domain-generalized embedding space that maintains robust class discriminability across varied known-novel class ratios. Traditional Task Arithmetic \cite{task_arithmetic} updates $\theta_{\text{global}}$ by integrating task vectors from fine-tuned models; however, applying TA directly for CD-GCD—using static task vectors derived from joint fine-tuning on $\mathcal{D}_{\mathcal{S}}$ and $\mathcal{D}_{\text{syn}}$—proves inadequate for handling dynamic distribution shifts and evolving semantics (Table \ref{tab:ablation}). Moreover, TA’s absence of validation-driven assessment increases the risk of overfitting to familiar domains, leading to diminished performance on unseen distributions, as reflected in {\ourmodel}\ (TA) results in Table \ref{tab:results}.

We introduce an episodic training framework with a novel dynamically weighted task-vector aggregation strategy to address these limitations (Fig.~\ref{fig:model1}). Training is structured across $n_g$ global updates, each containing $n_e$ episodes. At the start of the $e_g ^{th}$ episode, the task model is initialized with parameters $\theta_{\text{global}}^{g-1}$ from the previous global iteration. The model is then fine-tuned on the episode’s dataset pair $(\mathcal{D}_{\mathcal{S}}^{e_g}, \mathcal{D}_{\text{syn}}^{e_g})$ under the CD-GCD objective to obtain $\theta_{\text{local}}^{e_g}$. Subsequently, the task vector for the fine-tuned model, $\delta^{e}_g$, is computed by the parameter difference:

\begin{equation} 
\delta^{e}_g = \theta_{\text{global}}^{g-1} - \theta_{\text{local}}^{e_g} 
\label{eq:task_vector} 
\end{equation}

Rather than directly integrating $\delta^{e}_g$ into $\theta_{\text{global}}^{g-1}$, we propose weighting $\delta^{e}_g$ according to the GCD generalization ability of $\theta_{\text{local}}^{e_g}$, assessed on $\mathcal{D}_{\text{valid}}$. The \texttt{All} performance metric \cite{gcd}, which captures performance on both known and novel categories of $\mathcal{D}_{\text{valid}}$, serves as an effective measure of generalization; a high \texttt{All} score indicates superior generalizability. We further normalize the \texttt{All} scores across all episodes of the global update index $g$ (Fig. \ref{fig:model1}). Instead of using naive normalization methods like min-max normalization, we apply a \texttt{softmax} function to the \texttt{All} scores (see Table \ref{tab:ablation} for an comparative study). The \texttt{softmax} weighting emphasizes high-performing episodes by exponentially scaling weights, amplifying task vectors with stronger generalization potential while attenuating less effective ones. The generalization weight corresponding to $\theta_{\text{local}}^{e_g}$ is then calculated as follows:

\begin{equation} 
w_g^e = \frac{\exp(\texttt{all}^e_g)}{\sum_{e'=1}^{n_e} \exp(\texttt{all}^{e'}_g)} 
\label{eq:weighting} 
\end{equation}

 This adaptive weighting then informs the global update:

\begin{equation} 
\theta_{\text{global}}^g = \theta_{\text{global}}^{g-1} - \sum_{e=1}^{n_e} w_g^e \delta_g^e 
\label{eq:global_update} 
\end{equation}

Using a consistent validation set across episodes stabilizes task vector weighting, reduces gradient conflicts, and yields a more robust global model than baseline TA. Despite its resemblance with meta-learning \cite{vilalta2002perspective}, unlike traditional meta-learning, which focuses on rapid task adaptation, our framework centers on robust domain generalization, equipping the model to proactively manage evolving distributions. Further discussion on the \textit{meta-knowledge} acquired through our training process is available in Section \ref{sec:meta_knowledge}.\vspace{1mm}

\noindent \textbf{The CD-GCD objective within each episode, \ie $e_g^{th}$ episode:} 
Our CD-GCD objectives in each episode comprise three key loss functions: (i) a supervised contrastive loss for \(\mathcal{D}_{\mathcal{S}}^{e_g}\), (ii) an unsupervised contrastive objective for \(\mathcal{D}_{\mathcal{S}}^{e_g} \cup \mathcal{D}_{\text{syn}}^{e_g}\), and (iii) open-set domain adaptation coupled with a novel confidence optimization objective for fine-grained known and novel class samples in \(\mathcal{D}_{\mathcal{S}}^{e_g}\) and \(\mathcal{D}_{\text{syn}}^{e_g}\).

\textbf{- Supervised contrastive loss: } We utilize a supervised contrastive loss that operates in the embedding space of the encoder from the last global update, \(\mathcal{F}^{g-1}\), and aims to enhance the compatibility between an anchor sample \(x \in \mathcal{D}_{\mathcal{S}}^{e_g}\) and its corresponding class prototype (center) \(\mu^y\) obtained by averaging the embeddings of the class-wise samples, where \(y\) is the label of \(x\). Simultaneously, the loss reduces the similarity between the anchor and prototypes of all negative classes \( \mathcal{Y}_{s}^{e_g}\setminus y\). Given the temperature hyper-parameter $\tau$, $\mathcal{L}_{\text{con}}^s$ is given by:

\scriptsize
\begin{equation}
\mathcal{L}_{\text{con}}^s = - \mathbb{E}_{(x,y) \sim \mathcal{P}(\mathcal{D}_{\mathcal{S}}^{e_g})} \log \frac{\exp\left(\text{cos}(\mathcal{F}^{g-1}(x), \mu^y)/\tau\right)}{\sum_{y' \in \mathcal{Y}_{s}^{e_g}} \exp\left(\text{cos}(\mathcal{F}^{g-1}(x), \mu^{y'})/\tau\right)}
\end{equation}\normalsize

\textbf{- Unsupervised contrastive loss:} We employ an unsupervised contrastive loss to enable discriminative feature learning across domains. Each image $x$ and its geometrically augmented version $x^+$ (translation, rotation \etc) form a positive pair, while all other images in the batch $\mathcal{B}$, denoted $x^-$, serve as negatives. The objective is to maximize similarity between $x$ and $x^+$, while minimizing it between $x$ and all negatives $x^-$. This is formulated as:
\vspace{-2pt}

\scriptsize
\begin{equation}
\mathcal{L}_{\text{con}}^u =  \mathbb{E}_{x \sim \mathcal{P}(\mathcal{D}_{\mathcal{S}}^{e_g} \cup \mathcal{D}_{\text{syn}}^{e_g})} 
   -\log \frac{\exp\left( \operatorname{cos}(\mathcal{F}^{g-1}(x), \mathcal{F}^{g-1}(x^+)) / \tau \right)}
   {\sum\limits_{x' \in \mathcal{B}} \exp\left( \operatorname{cos}(\mathcal{F}^{g-1}(x), \mathcal{F}^{g-1}(x')) / \tau \right)}
\end{equation}
\normalsize

\begin{table*}[!ht]
  \centering
  \resizebox{0.9\textwidth}{!}{%
  \begin{tabular}{lccccc|ccc|ccc|cccccc}
    \toprule
    \multirow{2}{*}{\textbf{Methods}} & \multirow{2}{*}{\textbf{Venue}} & \textbf{Synthetic-} & \multicolumn{3}{c|}{\textbf{PACS}} & \multicolumn{3}{c|}{\textbf{Office-Home}} & \multicolumn{3}{c|}{\textbf{DomainNet}} & \multicolumn{3}{c}{\textbf{Average}} \\
    \cmidrule(lr){4-6} 
    \cmidrule(lr){7-9} 
    \cmidrule(lr){10-12} 
    \cmidrule(lr){13-15}
    & & \textbf{Domains} & \textbf{All} & \textbf{Old} & \textbf{New} & \textbf{All} & \textbf{Old} & \textbf{New} & \textbf{All} & \textbf{Old} & \textbf{New} & \textbf{All} & \textbf{Old} & \textbf{New} \\
    \midrule
    \cellcolor{pink!10}\textbf{ViT \cite{vit}} & \textit{ICLR'21} & \ding{55} & 41.98 & 50.91 & 33.16 & 26.17 & 29.13 & 21.62 & 25.35 & 26.48 & 22.41 & 31.17 & 35.51 & 25.73 \\
    \cellcolor{pink!10}\textbf{GCD \cite{gcd}} & \textit{CVPR'22} & \ding{55} & 52.28 & 62.20 & 38.39 & 52.71 & 54.19 & 50.29 & 27.41 & 27.88 & 26.13 & 44.13 & 48.09 & 38.27  \\
    \cellcolor{pink!10}\textbf{SimGCD \cite{gcd3}} & \textit{ICCV'23} & \ding{55} & 34.55 & 38.64 & 30.51 & 36.32 & 49.48 & 13.55 & 2.84 & 2.16 & 3.75 & 24.57 & 30.09 & 15.94  \\
    \cellcolor{pink!10}\textbf{CMS \cite{gcd6}} & \textit{CVPR'24} & \ding{55} & 28.95 & 28.13 & 36.80 & 10.02 & 9.66 & 10.53 & 2.33 & 2.40 & 2.17 & 13.77 & 13.40 & 16.50  \\
    \cellcolor{pink!10}\textbf{CDAD-Net \cite{rongali2024cdadnetbridgingdomaingaps}} & \textit{CVPR-W '24} & \ding{55} & 69.15 & 69.40 & 68.83 & 53.69 & \cellcolor{yellow!20}57.07 & 47.32 & 24.12 & 23.99 & 24.35 & 48.99 & 50.15 & 46.83  \\
    \cellcolor{pink!10}\textbf{\textsc{SODG-Net \cite{ODG3}}} & \textit{WACV'24} & \ding{55} & 37.43 & 40.28 & 28.38 & 36.53 & 49.42 & 14.58 & 27.77 & 27.98 & \cellcolor{yellow!20}26.15 & 33.91 & 39.23 & 23.04  \\
    \hline
    \cellcolor{gray!10}\textbf{GCD with Synthetic \cite{gcd}} & \textit{CVPR'22} & \ding{51} & 65.33 & 67.10 & 64.42 & 50.50 & 51.48 & 48.96 & 24.71 & 24.80 & 21.94  & 46.85 & 47.78 & 45.11  \\
    \cellcolor{gray!10}\textbf{SimGCD with Synthetic \cite{gcd3}} & \textit{ICCV'23} & \ding{51} & 39.76 & 43.76 & 35.97 & 35.57 & 48.58 & 12.89 & 2.71 & 1.99 & 4.14  & 26.01 & 31.44 & 17.67  \\
    \cellcolor{gray!10}\textbf{CMS with Synthetic \cite{gcd6}} & \textit{CVPR'24} & \ding{51} & 28.01 & 26.71 & 29.04 & 12.09 & 12.66 & 11.13 & 3.22 & 3.28 & 3.03  & 14.44 & 14.22 & 14.40 \\
    \cellcolor{gray!10}\textbf{CDAD-Net with Synthetic \cite{rongali2024cdadnetbridgingdomaingaps}} & \textit{CVPR-W '24} & \ding{51} & 60.76 & 61.67 & 59.49 & 53.49 & 56.90 & 47.76 & 23.85 & 23.88 & 24.26  & 46.03 & 47.47 & 43.84 \\
    \cellcolor{gray!10}{\textbf{DAML \cite{shu2021open}}} & \textit{CVPR '21} & \ding{51} & 40.26 & 42.90 & 29.28 & 36.20 & 49.53 & 14.06 & 27.10 & 28.25 & \cellcolor{red!18}26.16  & 34.52 & 40.23 & 23.17 \\
    \hline
    \cellcolor{green!5}\textbf{{{\ourmodel}} with TIES-Merging \cite{yadav2024ties}} & & \ding{51} & 67.04 & 71.25 & 64.02 & 53.52 & \cellcolor{red!18}57.12 & 48.03 & \cellcolor{yellow!20}28.72 & \cellcolor{yellow!20}30.35 & 24.39  & 49.76 & \cellcolor{yellow!20}52.91 & 45.48 \\
    \cellcolor{green!5}\textbf{{{\ourmodel}} with baseline TA\cite{task_arithmetic}} & & \ding{51} & 71.02 & 73.44 & 68.01 & 52.63 & 52.41 & 52.69 & 28.12 & 29.58 & 24.42  & 50.59 & 51.81 & 48.37 \\
    \hline
    \cellcolor{blue!10}\textbf{{{\ourmodel}} (Ours)} & & \ding{51} & \cellcolor{yellow!20}73.30 & \cellcolor{yellow!20}75.28 & \cellcolor{yellow!20}72.56 & \cellcolor{yellow!20}53.86 & 53.37 & \cellcolor{yellow!20}54.33 & \cellcolor{red!18}29.01 & \cellcolor{red!18}30.38 & 25.46  & \cellcolor{yellow!20}52.06 & \cellcolor{red!18}53.01 & \cellcolor{yellow!20}50.78 \\
    \cellcolor{blue!10}\textbf{{{\ourmodel} *} (Ours) [LoRA \cite{hu2021lora}]} & & \ding{51} & \cellcolor{red!18}75.21 & \cellcolor{red!18}76.31 & \cellcolor{red!18}75.28 & \cellcolor{red!18}54.32 & 53.03 & \cellcolor{red!18}56.03 & 27.17 & 27.61 & 25.99  & \cellcolor{red!18}52.23 & 52.32 & \cellcolor{red!18}52.43 \\
    \hline
    \textbf{CDAD-Net (DA) \cite{rongali2024cdadnetbridgingdomaingaps} [Upper bound]} & & \ding{55} & 83.25 & 87.58 & 77.35 & 67.55 & 72.42 & 63.44 & 70.28 & 76.46 & 65.19   & 73.69 & 78.82 & 68.66 \\
    \bottomrule
  \end{tabular}}
  \caption{\textbf{Performance comparison} across three datasets, averaged over all domain combinations. We baseline with two versions of existing GCD and CD-GCD methods: one trained solely on $\mathcal{D}_{\mathcal{S}}$ and another on $\mathcal{D}_{\mathcal{S}} \cup \mathcal{D}_{\text{syn}}$, both without target-specific loss functions to simulate the DG scenario. Subsequently, multiple model aggregation strategies are evaluated for {\ourmodel}. As an upper bound, we include the full CDAD-Net, designed for the transductive DA setting. Red cells highlight the top results; yellow is the second-best.}
  \label{tab:results}
  \vspace{-10px}
\end{table*}

 \textbf{- Domain-alignment objective with the proposed margin loss:} 
While the above objectives encourage learning discriminative features, effective domain alignment is crucial for transferring supervised knowledge from the labeled dataset $\mathcal{D}_{\mathcal{S}}^{e_g}$ to the unlabeled counterpart $\mathcal{D}_{\text{syn}}^{e_g}$. Given the label discrepancy between $\mathcal{D}_{\mathcal{S}}^{e_g}$ and $\mathcal{D}_{\text{syn}}^{e_g}$, we employ an open-set domain adaptation approach to avoid interference from non-overlapping classes of both the datasets. In contrast to the related literature, CDAD-Net \cite{rongali2024cdadnetbridgingdomaingaps}, which uses a closed-set adversarial loss for aligning the source and target domains but often fails to preserve the semantic structure of the novel classes (Table \ref{tab:ablation}), we purposefully push the novel classes away from the support of the known classes.

We build on the open-set adversarial domain adaptation objective, $\mathcal{L}_{\text{adv}}$, from \cite{saito2018open} for domain alignment. However, this $\mathcal{L}_{\text{adv}}$ often struggles with fine-grained separation between novel and known classes. To address this limitation, we introduce a margin objective $\mathcal{L}_{\text{margin}}$ that ensures each sample $x \in \mathcal{D}_{\text{syn}}^{e_g}$ is either confidently assigned to a known class or identified as an open-set sample with high probability, improving fine-grained differentiation between known and novel classes.
The combined domain alignment loss, $\mathcal{L}_{\text{da}}$, is defined as follows, where $\lambda$ is a weighting factor:

\begin{equation}
\centering
\mathcal{L}_{\text{da}} = \mathcal{L}_{\text{adv}} + \lambda \mathcal{L}_{\text{margin}}
\end{equation}

$\mathcal{L}_{\text{adv}}$ uses a $|\mathcal{Y}_{s}^{e_g}| + 1$-class episode-specific classifier $\mathcal{F}_{c}^{e_g}$ on top of \(\mathcal{F}^{g-1}\), with an additional class representing the open-set (details in Section \ref{sec:openset_recognition}). For $\mathcal{L}_{\text{margin}}$, let $p(x) = [p_1(x), \cdots, p_{|\mathcal{Y}_{s}^{e_g}|}(x)]$ denote predicted probabilities for known classes by $\mathcal{F}_{c}^{e_g}$ for $x$. We enforce a margin $m$ ($0 < m < 1$) between the highest known-class probability, $\max(p(x))$, and the open-set probability, $(1 - \sum_{q} p_q(x))$:

\scriptsize
\begin{equation}
\mathcal{L}_{\text{margin}} = \mathbb{E}_{x \in \mathcal{P}(\mathcal{D}_{\text{syn}}^{e_g})} 
\max \left( 0, \, m - \left| \max (p(x)) - \left( 1 - \sum_{q=1}^{|\mathcal{Y}_{s}^{e_g}|} p_q(x) \right) \right| \right)
\label{eq:margin_loss}
\end{equation}
\normalsize

$\mathcal{L}_{\text{margin}}$ enhances alignment between domains while preserving known-novel class distinctions.
The model is fine-tuned while \textbf{minimizing} \bm{$\mathbf{\mathcal{L}_{\text{con}}^s + \mathcal{L}_{\text{con}}^u + \mathcal{L}_{\text{da}}}$} \textbf{given} \bm{$\mathbf{[\mathcal{F}^{g-1}, \mathcal{F}_c^{e_g}}]$} to obtain $\theta_{\text{local}}^{e_g}$. Additionally, we examine resource-efficient fine-tuning using LoRA \cite{hu2021lora}, which reduces memory and computational requirements by adding trainable low-rank matrices to the existing weights. During inference, we employ the latest $\theta_{\text{global}}^{n_g}$ and apply clustering on samples from $\mathcal{D}_{\mathcal{T}}$. For the \textbf{pseudo-code}, see Section \ref{sec:pseudocode}

\section{Experimental Evaluations}\label{sec:experiments}

\label{sec:dataset}
\noindent \textbf{Datasets:} We evaluate the cross-domain capabilities of {\ourmodel}\ on three benchmark datasets commonly used for DG \cite{dg-survey1} and CD-GCD \cite{gcd}: (i) PACS \cite{li2017deeper} (4 domains, 9,991 samples, 7 classes), (ii) Office-Home \cite{officehome} (4 domains, 15,588 samples, 65 classes), and (iii) DomainNet \cite{peng2019moment} (6 domains, 586,575 samples, 345 classes). For PACS and Office-Home, each domain alternates as the source, with the remaining domains as targets. Source-target configurations for DomainNet are detailed in the Section \ref{sec:dataset_details}. The known-to-novel class ratios are 4:3 for PACS, 40:25 for Office-Home, and 250:95 for DomainNet, following \cite{rongali2024cdadnetbridgingdomaingaps}.

\noindent \textbf{Implementation details:} All methods employ the global model $\mathcal{F}$ with a ViT-B/16 backbone \cite{vit}, initialized with DINO pre-trained weights \cite{dino}, using the \texttt{[CLS]} token as the feature representation. Following the baseline in \cite{gcd}, only the final block of the vision transformer is fine-tuned in each local task model, with an initial learning rate of 0.1, using an SGD optimizer with cosine annealing decay. We set the batch size to 128 and $\lambda = 0.20$. We consider $n_g=10$, $n_e=6$, and each episode-specific model is fine-tuned for eight epochs. To balance separation between known and novel classes, we set the margin $m = 0.7$ in the margin loss $\mathcal{L}_{\text{margin}}$ (see Table \ref{tab:margin} for ablation). As an alternative, we also fine-tune local models using a LoRA variant with $\alpha = 16$ and \texttt{rank} = 8 for enhanced performance.

We adopt the \( K \)-estimation method described in \cite{gcd} for determining the optimal number of clusters \( K \). This involves applying Brent’s algorithm to features extracted from benchmark datasets, with \( K \) constrained between \(|\mathcal{Y}_s|\) and 1000 classes, ensuring efficiency and practicality.

\noindent \textbf{Evaluation:} Our primary evaluation metric, \texttt{All}, assesses clustering performance across the full \(\mathcal{D}_{\mathcal{T}}\). We also report results for two subsets: \texttt{Old}, containing instances from known classes (\(\mathcal{Y}_{t}^{old}\)), and \texttt{New}, containing instances from novel classes (\(\mathcal{Y}_{t}^{new}\)). Subset-specific accuracies are computed using the Hungarian optimal assignment algorithm \cite{kuhn1955hungarian} to determine the label permutation that maximizes accuracy. All reported results reflect the average performance across all domain combinations based on three runs.


\begin{figure*}[ht]
    \centering
    \includegraphics[width=0.8\linewidth]{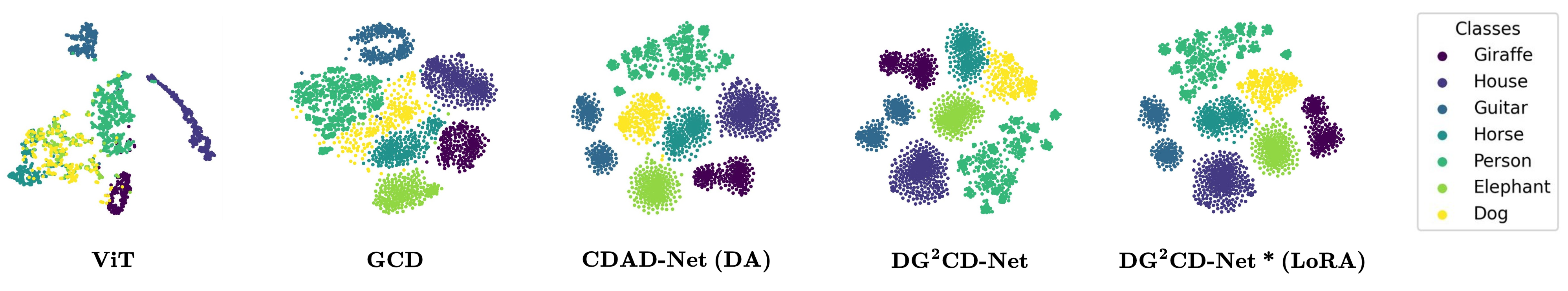}
    \vspace{-10px}
    \caption{\textbf{t-SNE \cite{van2008visualizing} visualizations} of the target domain (``Photo") clusters, as produced by pre-trained ViT, GCD \cite{gcd}, CDAD-Net(DA) \cite{rongali2024cdadnetbridgingdomaingaps}, DG$^{2}$CD-Net and DG$^{2}$CD-Net* (LoRA) for the PACS dataset, with ``Sketch" as the source domain. Both the variants of {\ourmodel}\ are able to produce a clean and compact embedding space.}
    \label{fig:tsne}
    \vspace{-15pt}
\end{figure*}

\subsection{Comparison to the literature}

We perform a comparative analysis of {{\ourmodel}} against DG and category discovery benchmarks. For single-source open-set DG, we include SODG-Net \cite{ODG3}. From the GCD literature, we evaluate ViT \cite{vit}, GCD \cite{gcd}, SimGCD \cite{gcd3}, CMS \cite{gcd6}, and CDAD-Net \cite{rongali2024cdadnetbridgingdomaingaps} as the representative of CD-GCD. All models are trained solely on source data, excluding target-specific loss functions to simulate DG-GCD. For GCD, SimGCD, CMS, and CDAD-Net, we also consider variants trained on the combined dataset $\mathcal{D}_{\mathcal{S}} \cup \mathcal{D}_{\text{syn}}$. Additionally, we include the multi-source open-set DG method DAML \cite{shu2021open}. CDAD-Net (DA) serves as an upper bound for domain adaptation, trained on both source and target domains. Results of CROW (DA) \cite{wen2024cross}, which can also serve as an upper bound, are mentioned in the Table \ref{tab:results_crow}.

Results on PACS, Office-Home, and DomainNet are presented in Table \ref{tab:results} (full results in Section \ref{sec:comparative_analysis}). The baseline ViT achieves an average \texttt{All} score of $31.17\%$. Among GCD and CD-GCD baselines, CDAD-Net (DG) achieves the highest performance (average \texttt{All}: $48.99\%$) when trained on $\mathcal{D}_{\mathcal{S}}$, demonstrating robustness in domain generalization without synthetic data. In contrast, vanilla GCD \cite{gcd} performs best (\texttt{All}: $46.85\%$) when trained on $\mathcal{D}_{\mathcal{S}} \cup \mathcal{D}_{\text{syn}}$, though the mixed distributions introduce gradient conflicts, affecting stability. CDAD-Net (DA) achieves an upper bound \texttt{All} score of $73.69\%$ in a transductive setting. SODG-Net and DAML trail behind {\ourmodel}\ by 18.15\% and 17.54\% in \texttt{All}, particularly struggling on the \texttt{New} metric due to ineffective clustering of novel classes.

{\ourmodel}\ achieves 49.76\% accuracy using only top-ranked model weights \cite{yadav2024ties}, which improves to 50.59\% with full-weight-space TA \cite{task_arithmetic}, suggesting that the full model space reduces performance ambiguity from gradient conflicts. Our episodic updates with validation-based weighting further boost performance to 52.06\%. With LoRA \cite{hu2021lora} replacing full fine-tuning, {\ourmodel}* reaches 52.23\%, indicating that domain-general knowledge resides in a compact manifold across datasets. t-SNE visualizations (Fig. \ref{fig:tsne}) show {\ourmodel}\ forming a compact embedding space.

\subsection{Estimating the number of clusters}
Table \ref{tab:cluster_results_comparison} highlights {\ourmodel}'s accuracy in estimating the number of clusters for PACS, Office-Home, and DomainNet datasets. {\ourmodel}\ aligns perfectly with the ground truth for PACS at 7 clusters and closely matches it for Office-Home and DomainNet with 67 and 355 clusters, respectively, outperforming CDAD-Net (DG) and achieving parity with CDAD-Net (DA). Such results illustrate our episodic training approach's strength in adapting to dynamic semantic shifts in $\mathcal{D}_{\mathcal{T}}$.

\begin{table}[!h]
\centering
\resizebox{0.7\linewidth}{!}{%
\begin{tabular}{cccc}
    \toprule
    \textbf{Dataset} & \textbf{PACS} & \textbf{Office-Home} & \textbf{DomainNet} \\ 
    \midrule
    \cellcolor{pink!10}\textbf{Ground Truth} & 7 & 65 & 345 \\
    \cellcolor{gray!5}\textbf{CDAD-Net (DG)} & 12 & 60 & 362 \\ 
    \cellcolor{gray!5}\textbf{CDAD-Net (DA)} & 7 & 66 & 349 \\ 
    \rowcolor{blue!10}
    \textbf{Ours} & 7 & 67 & 355 \\
    \bottomrule
\end{tabular}}
\vspace{-5px}
\caption{\textbf{Estimation of cluster numbers} in target domains showcases {\ourmodel}'s effectiveness in achieving near-optimal clustering, beating other DG counterparts, and closely matching CDAD-Net (DA) performance.}
\label{tab:cluster_results_comparison}
\vspace{-10pt}
\end{table}

\subsection{Main ablation study}

Besides the following experiments, additional ablation analyses concerning \textit{different splits of known-novel classes, different backbones, sensitivity to $m$ in Eq. \ref{eq:margin_loss}, model ablation on more datasets \etc} are mentioned in Section \ref{sec:ablations}.

\begin{figure}[ht]
    \centering
    \resizebox{0.9\linewidth}{!}{
    \begin{tikzpicture}
        \begin{axis}[
            width=9cm, 
            height=5cm,
            xlabel={Episodes},
            xlabel style={font=\footnotesize},
            ylabel={Accuracy (\%)},
            ylabel style={font=\footnotesize},
            xmin=0, xmax=12,
            ymin=30, ymax=80,
            xtick={0,2,6,8,12},
            ytick={30,40,50,60,70,80},
            legend style={
                font=\scriptsize,
                draw=black,
                fill=white,
                at={(1.05,1)}, 
                anchor=north west
            },
            grid=both,
            grid style={dashed, gray!30}
        ]
        \addplot[
            color=cvprblue,
            mark=*,
            ]
            coordinates {
            (0,50.91) (2,66.86) (6,73.3) (8,71.02) (12,69.1)
            };
        \addlegendentry{All}
        
        \addplot[
            color=cvprgreen,
            mark=square*,
            ]
            coordinates {
            (0,52.28) (2,69.11) (6,75.28) (8,73.44) (12,71.81)
            };
        \addlegendentry{Old}
        
        \addplot[
            color=red,
            mark=diamond*,
            ]
            coordinates {
            (0,33.16) (2,63.75) (6,72.56) (8,68.01) (12,65.31)
            };
        \addlegendentry{New}
        
        \end{axis}
    \end{tikzpicture}
    }
    \vspace{-10pt}
    \caption{Charting the relationship between \textbf{number of episodes and model accuracy} on PACS, this graph shows accuracy peaks at 6 episodes, which is majorly consistent across all the datasets.}
    \label{fig:episode}
   \vspace{-10pt}
\end{figure}
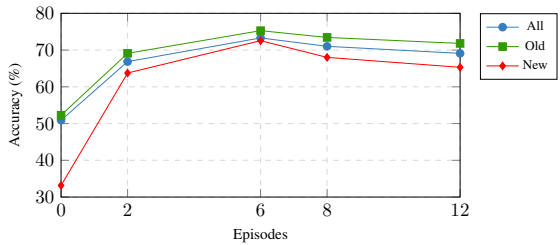

\noindent \textbf{(a) Impact of the number of episodes / synthetic domains in each global update:} We evaluate our model’s performance across varying numbers of training episodes, from 0 (using only the source domain) to 12, where each episode introduces a distinct synthetic domain. As shown in Fig. \ref{fig:episode}, training with up to 6 episodes significantly improves model accuracy, highlighting the impact of episodic learning for domain generalization. Beyond six episodes, however, gains diminish, likely due to increased sign conflicts—where opposing updates from different episodes lead to parameter divergence \cite{yadav2024ties}. This rise in sign conflicts, as visualized in Fig. \ref{fig:signconflicts}, correlates with the diminishing returns in accuracy, indicating a critical trade-off between generalization and stability. Optimizing the number of episodes is, therefore, vital to balance these effects.

\noindent \textbf{(b) Impact of model components:} We conducted experiments with several variants of our algorithm to underscore the impact of key components on domain generalization: (i) \textbf{Manual augmentations for} $\mathcal{D}_{\text{syn}}$: synthetic data generated through Gaussian blur, color jittering, and random rotations; (ii) \textbf{Without synthetic domain}: training solely on the source domain without target-specific loss (no CD-GCD locally); (iii) \textbf{Without multi-global updates}: using a single global update as in conventional Task Arithmetic (TA) \cite{task_arithmetic}; (iv) \textbf{Static known/novel splits}: fixing class splits across episodes to limit generalization over semantic shifts; (v) \textbf{Conventional normalization of $w_g^e$s}: using min-max instead of softmax normalization in Eq. \ref{eq:weighting}; (vi) \textbf{Episode-specific validation set} $\mathcal{D}_{\text{valid}}$: per-episode validation data ($\mathcal{D}_{\mathcal{S}}^{e_g} \cup \mathcal{D}_{\text{syn}}^{e_g}$) instead of a separate $\mathcal{D}_{\text{valid}}$; (vii) \textbf{Replacing losses with CDAD-Net} \cite{rongali2024cdadnetbridgingdomaingaps}: substituting CDAD-Net’s closed-set domain alignment loss for CD-GCD with ours; and (viii) \textbf{{\ourmodel}\ with Fisher-merging} \cite{matena2022merging}: incorporating Fisher-merging for model weighting. 

As shown in Table \ref{tab:ablation}, removing synthetic domains reduces the \texttt{All} metric by \(22.39\%\), while replacing them with manual augmentations decreases it by \(9.38\%\). A single global update instead of iterative updates lowers the \texttt{All} metric by \(6.60\%\), and static class splits lead to a \(4.70\%\) decrease in performance. Min-max normalization reduces \texttt{All} by \(11.32\%\), and episode-specific validation decreases it by \(4.04\%\), indicating the need for out-of-distribution validation data. Replacing our alignment and feature learning losses with CDAD-Net yields a \(3.48\%\) drop in \texttt{All}. Lastly, Fisher-merging inadequately captures domain variations, increasing gradient conflicts (Fig. \ref{fig:signconflicts}) and causing a \(1.63\%\) decline in \texttt{All}. These findings underscore the importance of synthetic domains, iterative updates, and our alignment techniques for robust domain generalization.

\begin{table}[h]
    \centering
    \vspace{-5px}
    \resizebox{0.9\linewidth}{!}{%
    \begin{tabular}{lccc}
        \toprule
        \multirow{2}{*}{\textbf{Model Variant}} & \multicolumn{3}{c}{\textbf{PACS}} \\
        \cline{2-4}
        & \textbf{All} & \textbf{Old} & \textbf{New} \\
        \midrule
        
         \textbf{(i) With manual augmentations based $\mathcal{D}_{\text{syn}}$} & 63.92 & 62.09 & 67.06 \\
         \textbf{(ii) Without synthetic domain} & 50.91 & 52.28 & 33.16 \\
        \textbf{(iii) Without multi-global updates} & 66.70 & 67.52 & 66.39 \\
         \textbf{(iv) Static known/novel class split across episodes} & 68.60 & 69.70 & 66.13 \\
         \textbf{(v) Conventional normalization for $w_g^e$s in Eq. \ref{eq:weighting}} & 61.98 & 69.77 & 54.43 \\
         \textbf{(vi) Episode specific $\mathcal{D}_{\text{valid}}$} & 69.26 & 73.34 & 65.15 \\
         \textbf{(vii) Replacing our CD-GCD losses with those of \cite{rongali2024cdadnetbridgingdomaingaps}} & 69.82 & 74.00 & 66.23 \\
         \textbf{(viii) {{\ourmodel}} with Fisher-merging \cite{matena2022merging}} & 71.67 & 74.42 & 70.03 \\
        \hline
        \rowcolor{blue!10}
        \textbf{  Full {{\ourmodel}}} & \textbf{73.30} & \textbf{75.28} & \textbf{72.56} \\
        \bottomrule
    \end{tabular}}
    \vspace{-5px}
    \caption{Performance metrics demonstrating the \textbf{influence of key model components} of {{\ourmodel}} for PACS.}
    \label{tab:ablation}
    \vspace{-12pt}
\end{table}

\noindent\textbf{(c) Impact of loss components:}
Table \ref{tab:performance_metrics} demonstrates the substantial improvements in classification accuracy on the PACS dataset achieved by systematically integrating loss components. Implementing $\mathcal{L}_{\text{con}}^{s}$ in \textbf{C-1} raised accuracy to 60.63\%, showcasing its efficacy with labeled data. With $\mathcal{L}_{\text{con}}^{u}$ in \textbf{C-2} accuracy boosted to 67.84\% through the use of unsupervised data. Incorporating $\mathcal{L}_{\text{adv}}$ in \textbf{C-6} elevated the accuracy to 71.45\%.  Finally, the complete integration in \textbf{C-7}, incorporating $\mathcal{L}_{\text{margin}}$, reached 73.30\%. These results underscore the synergistic effect of the combined loss functions, significantly enhancing the model’s generalization capability.

\begin{table}[!t]
    \centering
    \resizebox{0.8\linewidth}{!}{%
    \begin{tabular}{cccccccc}
        \toprule
        \multirow{2}{*}{\textbf{Configurations}} & \multirow{2}{*}{$\bm{\mathcal{L}_{con}^{s}}$} & \multirow{2}{*}{$\bm{\mathcal{L}_{con}^{u}}$} & \multirow{2}{*}{$\bm{\mathcal{L}_{adv}}$} & \multirow{2}{*}{$\bm{\mathcal{L}_{margin}}$} & \multicolumn{3}{c}{\textbf{PACS}} \\
        \cline{6-8}
         &  &  &  &  & \textbf{All} & \textbf{Old} & \textbf{New} \\
        \toprule
        \cellcolor{pink!10}\textbf{C-0} & \ding{55} & \ding{55} & \ding{55} & \ding{55} & 41.98 & 50.91 & 33.16 \\
        \cellcolor{red!10}\textbf{C-1} & \ding{51} & \ding{55} & \ding{55} & \ding{55} & 60.63 & 65.15 & 54.63 \\
        \cellcolor{red!10}\textbf{C-2} & \ding{55} & \ding{51} & \ding{55} & \ding{55} & 67.84 & 68.92 & 66.17 \\
        \cellcolor{gray!10}\textbf{C-3} & \ding{51} & \ding{51} & \ding{55} & \ding{55} & 68.70 & 69.52 & 67.61 \\
        \cellcolor{gray!10}\textbf{C-4} & \ding{51} & \ding{55} & \ding{51} & \ding{55} & 64.16 & 66.10 & 60.81 \\
        \cellcolor{gray!10}\textbf{C-5} & \ding{55} & \ding{51} & \ding{51} & \ding{55} & 69.22 & 69.20 & 69.34 \\
        \cellcolor{gray!10}\textbf{C-6} & \ding{51} & \ding{51} & \ding{51} & \ding{55} & 71.45 & 72.92 & 70.33 \\
        \rowcolor{blue!10}
        \textbf{C-7} & \ding{51} & \ding{51} & \ding{51} & \ding{51} & \textbf{73.30} & \textbf{75.28} & \textbf{72.56} \\
        \bottomrule
    \end{tabular}}
    \vspace{-5px}
    \caption{Impact of \textbf{loss components} of {{\ourmodel}} on PACS.}
    \label{tab:performance_metrics}
    \vspace{-15pt}
\end{table}

\subsection{Analyzing training convergence}

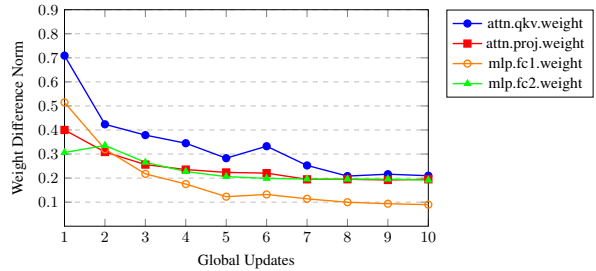
\begin{figure}[ht]
\vspace{-10px}
    \centering
    \resizebox{0.95\linewidth}{!}{
    \begin{tikzpicture}
        \begin{axis}[
            width=9cm,
            height=6cm,
            xlabel={Global Updates},
            xlabel style={font=\small},
            ylabel={Weight Difference Norm},
            ylabel style={font=\small},
            xmin=1, xmax=10,
            ymin=0, ymax=0.9,
            xtick={1,2,3,4,5,6,7,8,9,10},
            ytick={0.1,0.2,0.3,0.4,0.5,0.6,0.7,0.8,0.9},
            ymajorgrids=true,
            yminorgrids=true,
            grid style=dashed,
            legend style={
                font=\small,
                at={(1.05,1)},
                anchor=north west,
                draw=black,
                fill=white,
                align=left
            },
            cycle list name=color list,
            tick label style={font=\small},
        ]
        \addplot[thick, color=blue, mark=*] coordinates {
            (1,0.70921206) (2,0.42369723) (3,0.37868977) (4,0.34500438)
            (5,0.28279036) (6,0.3325212)  (7,0.25265133) (8,0.20843911)
            (9,0.21619427) (10,0.20989244)
        };
        \addlegendentry{attn.qkv.weight}

        \addplot[thick, color=red, mark=square*] coordinates {
            (1,0.400206)   (2,0.30854392) (3,0.25620466) (4,0.23543043)
            (5,0.22361384) (6,0.2206963)  (7,0.19489998) (8,0.19509061)
            (9,0.19225086) (10,0.1950625)
        };
        \addlegendentry{attn.proj.weight}

        \addplot[thick, color=orange, mark=o] coordinates {
            (1,0.51458687) (2,0.31995976) (3,0.21763138) (4,0.17529519)
            (5,0.12288372) (6,0.13203369) (7,0.11383786) (8,0.09969611)
            (9,0.09352239) (10,0.08964974)
        };
        \addlegendentry{mlp.fc1.weight}

        \addplot[thick, color=green, mark=triangle*] coordinates {
            (1,0.30630657) (2,0.33571145) (3,0.26554766) (4,0.22741044)
            (5,0.20612131) (6,0.19878145) (7,0.19622813) (8,0.19567198)
            (9,0.19595844) (10,0.19157822)
        };
        \addlegendentry{mlp.fc2.weight}
        \end{axis}
    \end{tikzpicture}
    }
    \vspace{-10px}
    \caption{Plot of the \textbf{average weight difference norm across global updates for different layers}, indicating model convergence and stabilization.}
    \label{fig:weight_diff}
    
\end{figure}

Fig. \ref{fig:weight_diff} illustrates the progressive weight difference norms for $\theta_{\text{global}}$ ($|\theta_{\text{global}}^g - \theta_{\text{global}}^{g-1}|_1$) across ten global updates in the attention projection weights and MLP layers of our ViT-B/16 block, trained on the ``Photo" domain of the PACS dataset. The results show a trend toward stabilized weight differences, indicating a smooth model convergence.

\vspace{-10px}

\section{Conclusions \& Future Directions}
\vspace{-5px}
This paper introduces DG-GCD, a novel GCD setting where the unlabeled target domain, distinct from the labeled source domain, is accessible only at test time. To address clustering under domain shifts without target-specific data, we propose {\ourmodel}, which integrates episodic training on source and synthetic pseudo-target domains with adaptive task-vector aggregation for a robust global model. Unlike traditional Task Arithmetic, our method dynamically weights task vectors based on the generalization performance of episode-specific models on held-out validation distributions. Additionally, an alignment objective with a novel margin loss ensures effective knowledge transfer, enabling confident differentiation of known and novel classes within episodes. Comprehensive benchmarks and ablation studies across multiple datasets validate the effectiveness of {\ourmodel}. We sincerely hope DG-GCD inspires further research in challenging areas like spotting new identities in person re-identification and novel drug discovery from chemical structures, where prior knowledge is scarce.

\noindent \textbf{Acknowledgment.} We acknowledge Adobe Research, CMInDS IITB for computational resources and our colleague Sona Elza Simon for insightful discussions.

\clearpage
\appendix
\onecolumn  
\begin{center}
    {\LARGE \textbf{Supplementary Material}}\\[1.5ex]
\end{center}

\section{What’s Inside This Supplementary Material?}

This supplementary material provides additional insights and extended analysis to support the main paper. Below, we outline the key sections :

\begin{description}
    \item[\textbf{Extended Literature on Class Discovery:}] (\hyperref[sec:lit_review]{Section~\ref*{sec:lit_review}}) A detailed review of prior work in Novel Category Discovery (NCD) and Generalized Category Discovery (GCD), emphasizing advancements relevant to domain shifts.
    \item[\textbf{DG-GCD in Practice: Applications Across Domains:}] (\hyperref[sec:dg_gcd_apps]{Section~\ref*{sec:dg_gcd_apps}}) Real-world applications of DG-GCD, including driverless cars, healthcare, and retail.
    \item[\textbf{Dataset Details:}] (\hyperref[sec:dataset_details]{Section~\ref*{sec:dataset_details}}) Comprehensive information on datasets, their configurations, and class distributions.
    \item[\textbf{Technical Details about Synthetic Data Generation:}] (\hyperref[sec:synthetic_data]{Section~\ref*{sec:synthetic_data}}) Description of the synthetic domain generation pipeline and parameter configurations.
    \item[\textbf{Synthetic Domain Utilization in DG-GCD:}] (\hyperref[sec:synthetic_utilization]{Section~\ref*{sec:synthetic_utilization}}) Integration of synthetic domains into training and validation, with visualizations and statistical insights.
    \item[\textbf{Handling Open-Set Recognition:}] (\hyperref[sec:openset_recognition]{Section~\ref*{sec:openset_recognition}}) Methodologies for open-set recognition, including episodic classifiers and adversarial loss.
    \item[\textbf{Pseudocode for Episodic Training Strategy:}] (\hyperref[sec:pseudocode]{Section~\ref*{sec:pseudocode}}) The pseudocode for our episodic training strategy and its implementation.
    \item[\textbf{Meta-Knowledge Learned in Episodic Training:}] (\hyperref[sec:meta_knowledge]{Section~\ref*{sec:meta_knowledge}}) An exploration of how meta-knowledge is acquired and refined through episodic training, enabling robust domain generalization and task adaptability.
    \item[\textbf{Comprehensive Comparative Analyses Across Datasets:}] (\hyperref[sec:comparative_analysis]{Section~\ref*{sec:comparative_analysis}}) Comparative analysis of {\ourmodel} on PACS, Office-Home, and DomainNet, evaluating its robustness and adaptability against benchmarks.
    \item[\textbf{Performance Comparison with Domain Adaptation Methods:}] (\hyperref[sec:performance_comparison]{Section~\ref*{sec:performance_comparison}}) A detailed comparison of {\ourmodel} with baseline and upper-bound domain adaptation methods.
    \item[\textbf{Additional Ablation Studies:}] (\hyperref[sec:ablations]{Section~\ref*{sec:ablations}}) An analysis of the contributions of individual components in {\ourmodel} across multiple datasets.
    \item[\textbf{Effect of Backbone Initialization:}] (\hyperref[sec:backbone_effect]{Section~\ref*{sec:backbone_effect}}) A comparison of results using different backbones.
    \item[\textbf{Effect of LoRA Fine-Tuning:}](\hyperref[sec:LORA_effect]{Section~\ref*{sec:LORA_effect}}) A comparison of results using {\ourmodel} with different LoRA adapters.
    \item[\textbf{Limitations and Future Work:}] (\hyperref[sec:limitations]{Section~\ref*{sec:limitations}}) A discussion on the current limitations and potential future directions for {\ourmodel}.
\end{description}

Each section is carefully crafted to provide deeper insights, reproducibility details, and additional context for the results presented in the main paper. We hope this supplementary material enhances the reader's understanding and offers a foundation for future exploration of domain generalization and category discovery.

\newpage

\section{Extended Literature on Class discovery} \label{sec:lit_review} Category discovery has evolved significantly from Novel Category Discovery (NCD) \cite{ncd1} to GCD \cite{gcd}. Traditionally, NCD methods \cite{ncd1, ncd2, ncd3, ncd4, ncd5} have utilized dual-model architectures where separate models are trained on labeled and unlabeled data to facilitate task transfer or employed parametric classifiers on top of generic feature extractors to categorize new classes.
Recent advancements in GCD focus on leveraging labeled data to generate pseudo-labels for unlabeled images. DCCL \cite{gcd1} uses InfoMap clustering, while PromptCAL \cite{gcd2} identifies pseudo-positive samples through semi-supervised affinity propagation techniques. PIM \cite{gcd4} improves this by optimizing bi-level mutual information, and SimGCD \cite{gcd3} utilizes knowledge distillation with a parametric classifier to enhance pseudo-label reliability. A key innovation is the contrastive mean-shift clustering from \cite{gcd6}, which creates a highly discriminative embedding space for fine-grained class distinctions. Additionally, Gaussian mixture models \cite{gcd7} have been explored for clustering in GCD.

\begin{table}[h]
    \centering
    \captionsetup{width=\columnwidth}
    \resizebox{0.6\linewidth}{!}{%
    \begin{tabular}{lccc}
        \toprule
        \multirow{2}{*}{\textbf{Methods}} & $\bm{\mathcal{D}_{\mathcal{T}}}$ \textbf{Available} & \textbf{Evaluation on} & \multirow{2}{*}{$\bm{\mathcal{P}(\mathcal{S}) = \mathcal{P}(\mathcal{T})}$} \\
        & \textbf{for Training} & \textbf{Known/Novel classes} & \\
        \midrule
        \textbf{NCD} \cite{hsu2017learning, ncd} & \ding{51} & \cellcolor{red!10}Novel & \ding{51} \\
        \textbf{GCD} \cite{gcd} & \ding{51} & \cellcolor{gray!10}Both & \ding{51} \\
        \textbf{CD-GCD} \cite{rongali2024cdadnetbridgingdomaingaps} & \ding{51} & \cellcolor{gray!10}Both & \ding{55} \\  \hline
        \rowcolor{blue!10}
        \textbf{DG-GCD} & \ding{55} & Both & \ding{55} \\
        \bottomrule
    \end{tabular}%
    }
    \caption{Comparison of methods based on target-domain data ($\mathcal{D}_{\mathcal{T}}$) availability during training, evaluation on known and/or novel classes, and distribution divergence between the source ($\mathcal{S}$) and target ($\mathcal{T}$) domains in class discovery contexts: novel (NCD), generalized (GCD), and cross-domain (CD-GCD). Our proposed DG-GCD setting is different and more challenging than the rest.}
    \label{tab:teaser}
\end{table}

\section{DG-GCD in Practice: Applications Across Domains} \label{sec:dg_gcd_apps}
The challenge of \textit{domain generalization for generalized category discovery (DG-GCD)} has numerous real-world applications across various industries. In \textit{autonomous vehicles}, models must adapt to changing environmental conditions (e.g., weather, lighting) and detect novel road objects without access to original training data. In \textit{healthcare}, medical image analysis must generalize across different diagnostic tools while discovering new pathologies, all while respecting patient privacy regulations. \textit{Wildlife monitoring} requires models to generalize across ecosystems while identifying new species, while \textit{surveillance systems} must detect novel threats and adapt to different environments such as airports and public spaces. In \textit{retail and e-commerce}, recommendation systems need to adapt across product categories and discover new items, while \textit{robotics} requires models to generalize in dynamic environments and detect new objects. Lastly, in \textit{precision agriculture}, systems must generalize across different farms and detect novel crop diseases without requiring access to original datasets. These applications highlight the importance of models that can generalize across domains and discover new categories in privacy-preserving, real-world scenarios.

\section{Dataset details} \label{sec:dataset_details}
\textbf{Datasets:} Our experiments are conducted on three benchmark datasets (i) \textbf{PACS} \cite{li2017deeper} (ii) \textbf{Office-Home} \cite{officehome}, and (iii) \textbf{DomainNet} \cite{peng2019moment}.

For PACS and Office-Home, each domain was used as the source, with all others as target domains. For DomainNet, a subset of source-target configurations was selected, as summarized in Table \ref{tab:domainnet_config}, ensuring the model was tested on diverse domain pairs.

\begin{table}[!h!t!p]
\centering
\footnotesize
\resizebox{0.5\columnwidth}{!}{%
\begin{tabular}{@{}c>{\centering\arraybackslash}p{6cm}@{}}
\toprule
\textbf{Source} & \textbf{Targets} \\ 
\midrule
Sketch & Clipart, Painting, Infograph, Quickdraw, Real World \\ 
Painting & Clipart, Sketch, Infograph, Quickdraw, Real World \\ 
Clipart & Painting, Sketch, Infograph, Quickdraw, Real World \\ 
\bottomrule
\end{tabular}}
\caption{Source-target configurations for DomainNet}
\label{tab:domainnet_config}
\end{table}

\begin{figure}[htbp]
    \centering
    \includegraphics[width=0.8\textwidth]{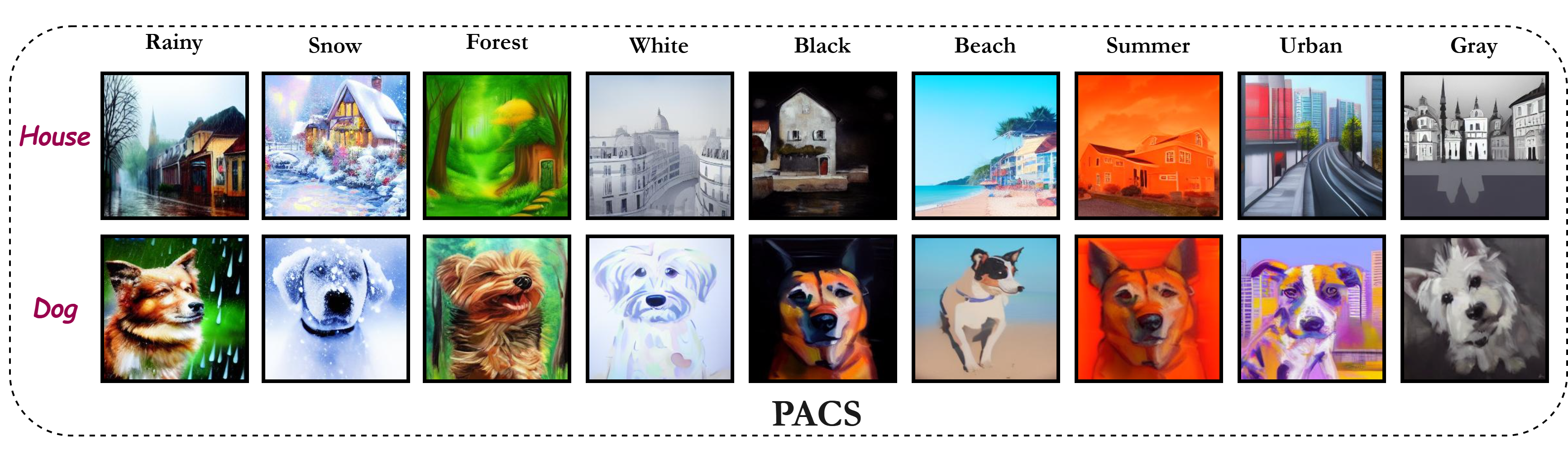}
    \caption{Additional synthetic image samples generated for the PACS dataset, showcasing diverse styles and domains across different categories (House and Dog).}
    \label{fig:sample}
\end{figure}

\section{Technical details about Synthetic Data Generation} \label{sec:synthetic_data}
We employed the InstructPix2Pix pipeline from Hugging Face's Diffusers library, utilizing the \texttt{timbrooks/instruct-} \texttt{pix2pix} model, for image transformation tasks. To achieve a balance between processing time and output quality, we configured the key parameter \texttt{num\_inference\_steps} to 10. The \texttt{image\_guidance\_scale} parameter was set to 1.0, ensuring that the model retained the essential structure of the input image while applying the specified transformations. Furthermore, the \texttt{guidance\_scale} parameter was adjusted to 7.5, promoting a strong alignment with the transformation prompt. These configurations allow for straightforward replication of our process while maintaining high-quality output.

\section{Synthetic Domain Utilization in DG-GCD} \label{sec:synthetic_utilization}

\begin{figure*}[!htbp]
    \centering
    \includegraphics[width=\textwidth]{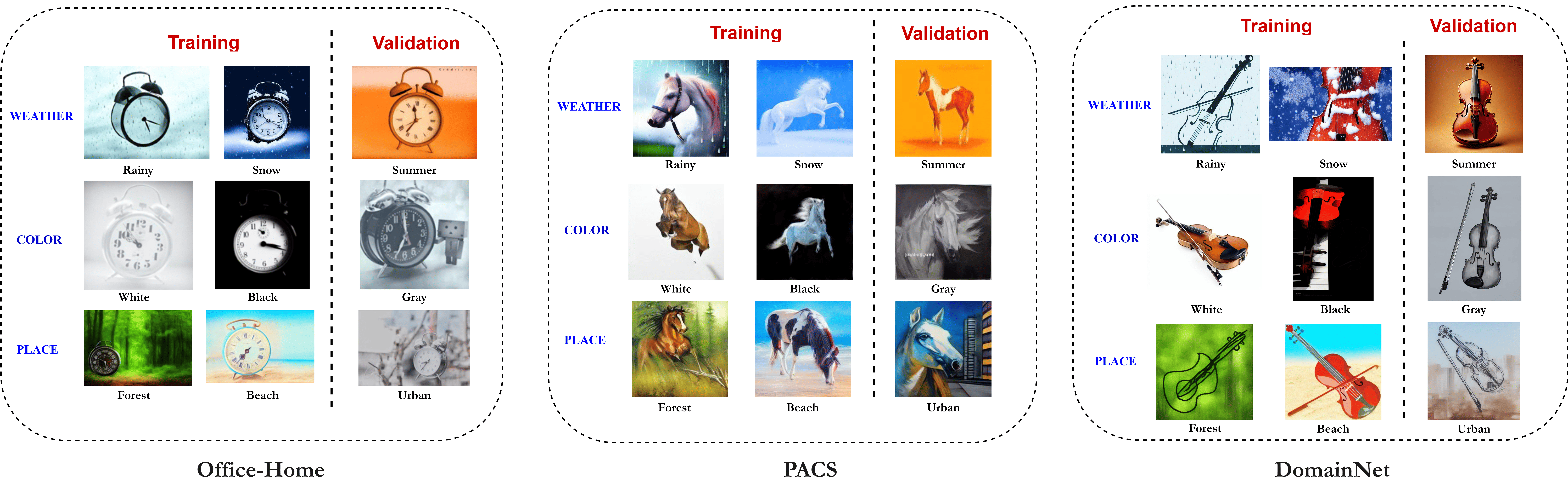}
    \caption{Categorization of synthetic domains utilized in the training and validation phases. Training domains are designed to simulate diverse conditions such as weather, color, and place variations. Validation domains challenge the model's adaptability to new, complex scenarios.}
    \label{fig:synthetic_domains}
\end{figure*}

In our study, we generate nine synthetic domains for each of the PACS, Office-Home, and DomainNet datasets to enhance the adaptability of models trained under the Domain Generalization for Generalized Category Discovery (DG-GCD) framework. Table \ref{tab:synthetic_domains} outlines the distribution of these domains, with six utilized in training and three in validation, ensuring comprehensive exposure to varied environmental conditions and rigorous testing of generalization capabilities. This structured approach improves the models' robustness against unseen real-world scenarios.

\begin{table}[!htbp] 
    \centering
    \setlength{\tabcolsep}{4pt} 
    \renewcommand{\arraystretch}{1.2} 
    
        \begin{tabular}{@{}lccc@{}}
            \toprule
            \textbf{Dataset} & \multicolumn{1}{p{2.5cm}}{\centering \textbf{Total Domains Generated}} & \multicolumn{1}{p{2.0cm}}{\centering \textbf{Used in Training}} & \multicolumn{1}{p{2.5cm}}{\centering \textbf{Used in Validation}} \\
            \midrule
            PACS         & 9 & 6 & 3 \\
            Office-Home  & 9 & 6 & 3 \\
            DomainNet    & 9 & 6 & 3 \\
            \bottomrule
        \end{tabular}
    
    \caption{Synthetic Domain Generation and Utilization for DG-GCD}
    \label{tab:synthetic_domains}
\end{table}

Figure \ref{fig:synthetic_domains} categorizes the synthetic domains used in our study's training and validation phases. The training domains include variations such as ``Rainy'' and ``Snow'' weather, ``White'' and ``Black'' colors, and ``Forest'' and ``Beach'' settings, broadening the model's exposure to diverse scenarios. The validation domains introduce ``Summer'' weather, ``Gray'' color, and ``Urban'' settings to test the model's ability to generalize across new and complex environments. This strategic use of synthetic domains demonstrates our approach to enhancing robustness and adaptability in models, crucial for effective domain generalization in real-world applications.

Table \ref{tab:frechet} presents the Average Fréchet Inception Distance (FID) scores for synthetic domains compared to the original domains within the Office-Home dataset. The scores are calculated to evaluate the visual similarity between generated images in environments like ``Rainy,'' ``Black,'' ``Urban,'' ``Beach,'' ``White,'' and ``Snow,'' and the original dataset categories: Art, Clipart, Product, and Real World. Lower FID scores indicate closer visual resemblance to the original domain, suggesting better synthetic image quality and domain adaptation. The data reveal variations in FID scores, with ``White'' and ``Black'' environments achieving the lowest scores, indicating higher similarity and potentially more effective domain generalization. This analysis provides insights into which synthetic modifications most accurately reflect the characteristics of their respective real-world counterparts, crucial for training robust models capable of generalizing across diverse visual contexts.

\begin{table}[!h!t!p]
    \centering
    \setlength{\tabcolsep}{4pt} 
    \renewcommand{\arraystretch}{1.2} 
    \begin{tabular}{l|cccc}
         \toprule
         \multirow{2}{*}{\textbf{Domain}} & \multicolumn{4}{c}{\textbf{FID Scores}} \\ \cline{2-5}
         & \textbf{Art} & \textbf{Clipart} & \textbf{Product} & \textbf{Real World} \\
         \midrule
         \textbf{Rainy}  & 140.11 & 158.81 & 167.27 & 154.42 \\
         \textbf{Black}  & 58.69  & 74.35  & 86.65  & 72.28  \\ 
         \textbf{Urban}  & 97.77  & 131.80 & 138.45 & 119.78 \\
         \textbf{Beach}  & 109.13 & 133.33 & 136.54 & 121.10 \\
         \textbf{White}  & 49.80  & 71.07  & 77.34  & 61.73  \\
         \textbf{Snow}   & 100.99 & 125.03 & 132.42 & 117.50 \\
         \bottomrule
    \end{tabular}
    
    \caption{Average Fréchet Inception Distance (FID) comparison between pairs of the generated and original domains over all the classes of the Office-Home dataset.}
    \label{tab:frechet}
\end{table}

\section{Handling Open-Set Recognition with Episode-Specific Classifiers} \label{sec:openset_recognition}

In our framework, we employ a \(|\mathcal{Y}^{e_g}_s|+1\)-class episode-specific classifier to address open-set recognition in episodic training. Here, \(|\mathcal{Y}^{e_g}_s|\) denotes the number of known classes in an episode, while the additional class models the open-set category, capturing instances outside the known classes. This design dynamically adapts to episodic data, ensuring robust differentiation between known and unknown classes.

\subsection{Adversarial Loss in Open-Set Domain Adaptation}

To refine the decision boundary, we use an adversarial loss, \(\mathcal{L}_{\text{adv}}\), which separates open-set instances by pushing them further from the known classes in feature space. This adversarial refinement enables the confident classification of unknown samples while maintaining performance on known categories. By tailoring decision boundaries to the episodic data, the classifier improves adaptive learning, generalization to new domains, and detection of novel classes in unseen distributions. 

The adversarial loss used in our method is inspired by \cite{saito2018open}. It facilitates the separation of known and unknown samples in the target domain by training the classifier $\mathcal{F}_{c}^{e_g}$ and the generator \(\mathcal{F}^{g-1}\) adversarially. The adversarial loss \(\mathcal{L}_{\text{adv}}\) is defined as:
\begin{equation}
    \mathcal{L}_{\text{adv}}(x^t) = - \alpha \log(p(y = |\mathcal{Y}^{e_g}_s|+1 | x^t)) - (1 - \alpha) \log(1 - p(y = |\mathcal{Y}^{e_g}_s|+1 | x^t)),
\end{equation}
Where:
\begin{itemize}
    \item \(x^t\) represents a target sample from $\mathcal{D_{\text{syn}}}$,
    \item \(p(y = |\mathcal{Y}^{e_g}_s|+1 | x^t)\) is the predicted probability that \(x^t\) belongs to the unknown class,
    \item \(\alpha\) is a hyperparameter (set to 0.5 in our experiments) that determines the decision boundary for the unknown class.
\end{itemize}

\subsubsection{Training Objectives}

Adversarial loss ($\mathcal{L}_{adv}$) is optimized along with Source Classification loss, which is a standard cross-entropy loss that ensures accurate classification of known source samples:
\begin{equation}
        \mathcal{L}_s(x^s, y^s) = -\log(p(y = y^s | x^s)),
\end{equation}
where \((x^s, y^s)\) represents a source sample and its label  from $\mathcal{D_{\mathcal{S}}}$ respectively.
    
The classifier $\mathcal{F}_{c}^{e_g}$ and generator \(\mathcal{F}^{g-1}\)  is trained in the following manner:

\begin{itemize}
        \item For the classifier $\mathcal{F}_{c}^{e_g}$, the objective is to minimize the total loss:
        \begin{equation}
            \min_{\mathcal{F}_{c}^{e_g}} \mathcal{L}_s(x^s, y^s) + \mathcal{L}_{\text{adv}}(x^t).
        \end{equation}
        \item For the generator \(\mathcal{F}^{g-1}\), the objective is to deceive the classifier by maximizing the adversarial loss:
        \begin{equation}
            \min_{\mathcal{F}^{g-1}} \mathcal{L}_s(x^s, y^s) - \mathcal{L}_{\text{adv}}(x^t).
        \end{equation}
\end{itemize}

\subsubsection{Implementation Details}
To efficiently compute the adversarial loss, we use a Gradient Reversal Layer (GRL), which flips the gradient sign during backpropagation. This allows simultaneous updates to $\mathcal{F}_{c}^{e_g}$ and \(\mathcal{F}^{g-1}\), facilitating stable training of adversarial objectives.


\section{Pseudocode of the proposed Episodic Training Strategy} \label{sec:pseudocode}


In this section, we present the pseudocode for the proposed episodic training strategy, as detailed in the main paper. This algorithm is used to iteratively update the global model parameters $\theta_{\text{global}}$ across multiple episodes and global updates. The process involves training task models on synthetic domains and updating the global model based on task vector computations and validation results.

\begin{algorithm}
\caption{Proposed Episodic Training Strategy for Updating $\theta_{\text{global}}$}
\label{alg:global_update}
\begin{algorithmic}[1] 
    \Require Pre-trained global model parameters $\theta_{\text{global}}^{0}$, number of global updates $n_g$, number of episodes per global update $n_e$, source domain $\mathcal{D}_{\mathcal{S}}^{e_{g}}$, synthetic domain $\mathcal{D}_{\text{syn}}^{e_{g}}$, validation domain $\mathcal{D}_{\text{valid}}$ in the $e_g^{th}$ episode.
    \Ensure Final global model $\theta_{\text{global}}^{n_g}$.
    
    \For{$g = 1$ to $n_g$} \Comment{Global updates}
        \State Randomly shuffle synthetic domains $\mathcal{D}_{\text{syn}}^{e_{g}}$.
        \For{each episode $e = 1$ to $n_e$} \Comment{Episode training}
            \State Initialize task model parameters $\theta_{\text{local}}^{e_g} \gets \theta_{\text{global}}^{g-1}$.
            \State Train task model $\theta_{\text{local}}^{e_g}$ on $(\mathcal{D}_{\mathcal{S}}^{e_{g}}, \mathcal{D}_{\text{syn}}^{e_{g}})$ for the CD-GCD task.
            \State Compute task vector $\delta^e_g$ (Equ. 1 in the main text) :
            \[
            \delta^{e}_g = \theta_{\text{global}}^{g-1} - \theta_{\text{local}}^{e_g}
            \]
            \State Validate task model on $\mathcal{D}_{\text{valid}}$ and compute accuracies: \texttt{All}, \texttt{Old}, \texttt{New}.
        \EndFor
        \State Compute weights $w_g^e$ using softmax on \texttt{All} accuracies (Equ. 2 in main text) :
        \[
        w_g^e = \frac{\exp(\texttt{All}^e_g)}{\sum_{e'=1}^{n_e} \exp(\texttt{All}^{e'}_g)}
        \]
        \State Update global model $\theta_{\text{global}}^g$ (Equ. 3 in main text) :
        \[
        \theta_{\text{global}}^g = \theta_{\text{global}}^{g-1} - \sum_{e=1}^{n_e} w_g^e \delta_g^e
        \]
    \EndFor
    \State Save the final global model: $\theta_{\text{global}}^{n_g}$.
\end{algorithmic}
\end{algorithm}

\section{Meta-Knowledge learnt in Episodic Training} \label{sec:meta_knowledge}
In our episodic training framework, meta-knowledge encompasses the cumulative insights gained from dynamic adaptation to varying domain conditions. This knowledge is perpetually refined via systematic application and iterative adjustment of task vectors, informed by feedback from domain-specific fine-tuning and rigorous validation processes. Unlike conventional meta-learning, which primarily targets rapid task adaptability, our framework emphasizes robust domain generalization. This approach enhances the model's proficiency in effectively preemptively addressing and adapting to evolving data distributions.

Meta-knowledge is acquired through:
\begin{itemize}
    \renewcommand\labelitemi{[-]}
    \item \textbf{Cross-Domain Exposure:} By engaging with overlapping features across multiple domains, the model develops a nuanced capability to generalize across diverse training environments. This cross-domain learning is fundamental in enabling the model to abstract and apply domain-invariant patterns to new, unseen scenarios.
    \item \textbf{Dynamic Vector Adjustments:} Task vectors are continually updated in response to real-time performance metrics. This dynamic refinement process allows the model to adjust its generalization strategies on the fly, enhancing its responsiveness to changes in domain characteristics.
    \item \textbf{Validation-Driven Learning:} The integration of internal validation mechanisms ensures continuous performance feedback. This feedback is instrumental in fine-tuning the model's strategic adjustments, ensuring optimized responses to future domain shifts and data interactions.
\end{itemize}

Our method enhances the generalizability of pre-trained foundation models by adaptively combining their fine-tuned versions across multiple automatically synthesized domains, eliminating the need for manual annotations. This adaptive strategy leverages the generalization performance of each fine-tuned model, minimizing the impact of poorly generalizable models. Besides, our approach ensures both discriminativeness and domain independence for the DG-GCD task. Consequently, this produces an embedding space predominantly guided by class semantics and suppressing stylistic artifacts, making it highly effective for generalization and clustering.

\section{More details on comparisons to literature} 

\subsection{DG-GCD specific adaptations for baselines}

We evaluate several state-of-the-art methods for generalized category discovery (GCD) and domain generalization (DG) by adapting them to the DG-GCD setting, as detailed in Table 2 of the main text. In this setting, target domain access is completely removed during training, and for certain methods, synthetic domain data is incorporated to simulate domain shifts.

For ViT-B/16, pre-trained with DINO, we fine-tuned only the last block using source domain data, following standard GCD practices, and evaluated the model on target domains without any target domain data during training. Similarly, GCD was adapted by fine-tuning the last block of the backbone on the source domain alone, and we introduced a synthetic domain variant to account for domain shifts.

For CMS (Contrastive Mean Shift) and SimGCD, we followed a similar procedure. We fine-tuned the last block using only the source domain and, in addition, created synthetic variants by incorporating synthetic domain data to assess the methods’ ability to handle domain shifts effectively.

CDAD-Net, which is designed for cross-domain adaptation, was also adapted for the DG-GCD setting. We ensured that it trained solely on the source domain, without access to target domain data, and created a synthetic variant to evaluate its performance on unseen domains.

As seen in Table 2 of the main text, the methods incorporating synthetic domain data, such as GCD with synthetic data, generally performed better in handling domain shifts, especially for novel class discovery. Our proposed models, leveraging task merging techniques such as TIES-Arithmetic and LoRA, outperform baseline methods on both benchmark datasets, achieving superior results, particularly on novel classes. The inclusion of synthetic domains proves beneficial, as evidenced by the marked performance improvement across all datasets, with our model consistently achieving the highest or second-highest results.

\subsection{Comprehensive comparative analyses of {\ourmodel} across multiple datasets} \label{sec:comparative_analysis}

This section presents a concise comparative analysis of {\ourmodel} on PACS, Office-Home, and DomainNet in Table-\ref{tab:PACS_comparison2}, \ref{tab:office_home_comparison2} and \ref{tab:domain_net_comparison3} respectively. Each dataset challenges {\ourmodel} with unique domain shifts, showcasing its adaptability and robustness. This evaluation aims to validate {\ourmodel}'s performance against established benchmarks, highlighting it's strengths and identifying opportunities for advancement in domain generalization.


\begin{table}[!htbp]
\centering
\resizebox{0.98\textwidth}{!}{%
\begin{tabular}{>{\bfseries}lccc|ccc|ccc|ccc|ccc|ccc}
\toprule
\multicolumn{19}{c}{\textbf{PACS}} \\ 
\midrule
 \multirow{2}{*}{\textbf{Methods}} & \multicolumn{3}{c|}{\textbf{Art Painting $\rightarrow$ Sketch}} & \multicolumn{3}{c|}{\textbf{Art Painting $\rightarrow$ Cartoon}} & \multicolumn{3}{c|}{\textbf{Art Painting $\rightarrow$ Photo}} & \multicolumn{3}{c|}{\textbf{Photo $\rightarrow$ Art Painting}} & \multicolumn{3}{c|}{\textbf{Photo $\rightarrow$ Cartoon}} & \multicolumn{3}{c}{\textbf{Photo $\rightarrow$ Sketch}} \\
 \cline{2-4} \cline{5-7} \cline{8-10} \cline{11-13} \cline{14-16} \cline{17-19}
 & \texttt{All} & \texttt{Old} & \texttt{New} & \texttt{All} & \texttt{Old} & \texttt{New} & \texttt{All} & \texttt{Old} & \texttt{New} & \texttt{All} & \texttt{Old} & \texttt{New} & \texttt{All} & \texttt{Old} & \texttt{New} & \texttt{All} & \texttt{Old} & \texttt{New} \\
\midrule
ViT \cite{vit}                  & 37.44                            & \cellcolor{yellow!20}50.73        & 19.5         & 47.4                              & 61.3         & 35.25        & 76.05                           & 87.13        & 64.64        & 53.17                           & 77.31        & 31.67        & 47.01                     & 55.54        & 39.57        & 31.87                    & 37.57        & 24.16        \\
GCD \cite{gcd}                  & 32.02                            & 41.53        & 19.12        & 46.78                             & 60.35        & 28.57        & 79.16                           & 99.45        & 48.73        & 74.73                           & 80.26        & 67.31        & 57.53                     & 60.46        & 53.6         & 46.23                    & \cellcolor{yellow!20}48.56        & 43.08        \\
SimGCD \cite{gcd3}               & 29.35                            & 17.3         & \cellcolor{red!18}62.12        & 23.08                             & 28.26        & 16.32        & 51.98                           & 74.44        & 33.26        & 46.29                           & 48.96        & 43.17        & 34.26                     & 44.91        & 20.35        & 24.84                    & 31.88        & 5.68         \\
CDAD-Net \cite{rongali2024cdadnetbridgingdomaingaps}                 & 46.02                            & 45.95        & 46.21        & 51.71                             & 53.43        & \cellcolor{yellow!20}49.46        & 99.04                           & 99.21        & 98.9         & 76.61                           & 76.97        & 76.19        & 56.78                     & 56.67        & 56.93        & \cellcolor{yellow!20}46.65                    & 46.15        & 48.01        \\
GCD With Synthetic   & 45.78                            & 36.71        & 58.01        & 54.84                             & \cellcolor{red!18}73.47        & 38.57        & 82.6                            & 66.29        & \cellcolor{yellow!20}99.39        & 79                              & 86.84        & 72.02        & 53.56                     & 67.93        & 41.01        & 44.18                    & 47.78        & 39.32        \\
CDAD-Net with Synthetic  & 43.09                            & 42.53        & 44.6         & 49.45                             & 59.31        & 36.58        & 99.16                           & 99.21        & 99.12        & 65.38                           & 62.83        & 68.36        & 42.92                     & 41.97        & 44.15        & 41.51                    & 43.79        & 35.32        \\
{\ourmodel}(TIES-Merging\cite{yadav2024ties}) & 41.31                            & 40.56        & 42.31        & 45.69                             & 57           & 35.81        & 96.11                           & 97.87        & 94.29        & 62.87                           & 87.72        & 40.72        & 48.98                     & 60.02        & 39.33        & 44.1                     & 36.33        & 54.58        \\
{\ourmodel}[TA\cite{task_arithmetic}]           & 46.4                             & \cellcolor{red!18}51.3         & 42.8         & 56.21                             & 58.82        & 52.7         & 99.2                            & \cellcolor{yellow!20}99.5         & 98.8         & 81.47                           & 91.09        & 68.57        & 57.76                     & 56.11        & \cellcolor{red!18}59.99        & \cellcolor{red!18}46.88                    & \cellcolor{red!18}48.96        & 44.06        \\
{\ourmodel}(Ours)         & \cellcolor{yellow!20}46.79                            & 38.13        & 58.49        & \cellcolor{yellow!20}57.96                             & \cellcolor{yellow!20}73.38        & 44.48        & \cellcolor{yellow!20}99.34                           & \cellcolor{red!18}99.7         & 98.97        & \cellcolor{yellow!20}86.67                           & \cellcolor{yellow!20}91.87        & \cellcolor{yellow!20}82.04        & \cellcolor{yellow!20}62.97                     & \cellcolor{yellow!20}71.18        & 55.8         & 45.72                    & 36.53        & \cellcolor{yellow!20}58.13        \\
{\ourmodel}* (Ours)[LoRA\cite{hu2021lora}]   & \cellcolor{red!18}46.83                            & 37.79        & \cellcolor{yellow!20}59.03        & \cellcolor{red!18}63.82                             & 71.13        & \cellcolor{red!18}57.43        & \cellcolor{red!18}99.46                           & 99.35        & \cellcolor{red!18}99.57        & \cellcolor{red!18}88.89                           & \cellcolor{red!18}93.94        & \cellcolor{red!18}84.4         & \cellcolor{red!18}64.19                     & \cellcolor{red!18}72.23        & \cellcolor{yellow!20}57.15        & 46.45                    & 37.75        & \cellcolor{red!18}58.19  \\ 
\bottomrule
\end{tabular}}

\vspace{1px}

\resizebox{0.98\textwidth}{!}{%
\begin{tabular}{>{\bfseries}lccc|ccc|ccc|ccc|ccc|ccc}
\midrule
 \multirow{2}{*}{\textbf{Methods}} & \multicolumn{3}{c|}{\textbf{Sketch $\rightarrow$ Art Painting}} & \multicolumn{3}{c|}{\textbf{Sketch $\rightarrow$ Cartoon}} & \multicolumn{3}{c|}{\textbf{Sketch $\rightarrow$ Photo}} & \multicolumn{3}{c|}{\textbf{Cartoon $\rightarrow$ Art Painting}} & \multicolumn{3}{c|}{\textbf{Cartoon $\rightarrow$ Sketch}} & \multicolumn{3}{c}{\textbf{Cartoon $\rightarrow$ Photo}} \\
 \cline{2-4} \cline{5-7} \cline{8-10} \cline{11-13} \cline{14-16} \cline{17-19}
 & \texttt{All} & \texttt{Old} & \texttt{New} & \texttt{All} & \texttt{Old} & \texttt{New} & \texttt{All} & \texttt{Old} & \texttt{New} & \texttt{All} & \texttt{Old} & \texttt{New} & \texttt{All} & \texttt{Old} & \texttt{New} & \texttt{All} & \texttt{Old} & \texttt{New} \\
\midrule
ViT \cite{vit}                  & 23.93                 & 26.53    & 21.61    & 40.61           & 58.92    & 24.62    & 33.29        & 33.88    & 32.69    & 38.09                  & 47.36 & 29.82 & 33.57           & 35.67 & 30.74 & 41.38         & 39.08 & 43.74  \\
GCD \cite{gcd}                  & 33.25                 & 39.09    & 25.43    & 40.89           & 48.14    & 31.17    & 46.86        & 59.28    & 28.22    & 58.15                  & 78.52 & 30.86 & 36              & \cellcolor{yellow!20}44.83 & 24.04 & 75.75         & 85.88 & 60.55   \\
SimGCD \cite{gcd3}               & 21.19                 & 31.91    & 8.67     & 23.17           & 36.77    & 5.4      & 34.22        & 27.46    & 40.8     & 38.38                  & 42.07 & 34.07 & 34.84           & 33.94 & 37.31 & 53.05         & 45.85 & 59.06  \\
CDAD-Net \cite{rongali2024cdadnetbridgingdomaingaps}                 & 87.99                 & 84.32    & \cellcolor{red!18}92.28    & 51.88           & 51.77    & 52.02    & 99.04        & 99.21    & 98.9     & 73.05                  & 76.88 & 68.57 & 41.84           & 42.71 & 39.49 & 99.22         & 99.47 & 99.01   \\
GCD With Synthetic~  & 82.15                 & 85.13    & 79.5     & 44.3            & 48.22    & 40.89    & \cellcolor{red!18}99.49        & \cellcolor{red!18}99.76    & 99.21    & 63.01                  & 63.73 & 62.37 & 35.66           & 29.95 & 43.36 & \cellcolor{red!18}99.43         & \cellcolor{red!18}99.47 & 99.39  \\
CDAD-Net with Synthetic  & 61.91                 & 69.45    & 53.12    & 48.59           & 53.13    & 42.67    & 68.44        & 63.5     & 72.56    & 67.24                  & 65.28 & 69.52 & 42.05           & 39.61 & 48.67 & \cellcolor{yellow!20}99.34         & \cellcolor{yellow!20}99.47 & 99.23   \\
{\ourmodel}(TIES-Merging\cite{yadav2024ties}) & 80.59              & 80.78 & 80.42  & \cellcolor{yellow!20}58.94        & \cellcolor{yellow!20}75.71  & 44.28  & 99.07     & 98.64 & 99.51  & 87.45                  & 90.93 & 84.35 & 40.67           & 31.39 & 53.2  & 98.71         & 97.99 & 99.45  \\
{\ourmodel}(TA\cite{task_arithmetic})           & 73.02                 & 79.37    & 64.51    & 55.89           & 54.84    & \cellcolor{red!18}57.29    & \cellcolor{yellow!20}99.31        & \cellcolor{yellow!20}99.5     & 99.03    & \cellcolor{yellow!20}90.89                  & 92.75 & \cellcolor{yellow!20}88.4  & 46.03           & \cellcolor{red!18}49.67 & 41.1  & 99.16         & 99.35 & 98.88  \\
{\ourmodel}(Ours)         & \cellcolor{yellow!20}88.75                 & \cellcolor{yellow!20}93.52    & 84.49    & 56.76           & 72.14    & 43.33    & 99.13        & 98.7     & \cellcolor{yellow!20}99.57    & 90.77                  & \cellcolor{yellow!20}93.37 & \cellcolor{red!18}88.46 & \cellcolor{red!18}49.2            & 43.18 & \cellcolor{yellow!20}57.33 & 95.57         & 91.62 & \cellcolor{yellow!20}99.64  \\
{\ourmodel}* (Ours)[LoRA\cite{hu2021lora}]   & \cellcolor{red!18}90.87                 & \cellcolor{red!18}95.28    & \cellcolor{yellow!20}86.93    & \cellcolor{red!18}66.25           & \cellcolor{red!18}78.32    & \cellcolor{yellow!20}55.72    & 99.22        & 98.88    & \cellcolor{red!18}99.57    & \cellcolor{red!18}91.02                  & \cellcolor{red!18}93.99 & 88.37 & \cellcolor{yellow!20}46.33           & 38.19 & \cellcolor{red!18}57.33 & 99.22         & 98.82 & \cellcolor{red!18}99.64 \\
\bottomrule
\end{tabular}}
\caption{Detailed comparison of our proposed {\ourmodel} on DG-GCD with respect to referred literature for PACS Dataset.}
\label{tab:PACS_comparison2}
\end{table}

\begin{table}[!htbp]
\centering
\resizebox{0.98\textwidth}{!}{%
\begin{tabular}{>{\bfseries}lccc|ccc|ccc|ccc|ccc|ccc}
\toprule
 \multicolumn{19}{c}{\textbf{Office-Home}} \\ 
\midrule
 \multirow{2}{*}{\textbf{Methods}} & \multicolumn{3}{c|}{\textbf{Art $\rightarrow$ Clipart}} & \multicolumn{3}{c|}{\textbf{Art $\rightarrow$ Product}} & \multicolumn{3}{c|}{\textbf{Art $\rightarrow$ Real World}} & \multicolumn{3}{c|}{\textbf{Clipart $\rightarrow$ Art}} & \multicolumn{3}{c|}{\textbf{Clipart $\rightarrow$ Real World}} & \multicolumn{3}{c}{\textbf{Clipart $\rightarrow$ Product}} \\
 \cline{2-4} \cline{5-7} \cline{8-10} \cline{11-13} \cline{14-16} \cline{17-19}
 & \texttt{All} & \texttt{Old} & \texttt{New} & \texttt{All} & \texttt{Old} & \texttt{New} & \texttt{All} & \texttt{Old} & \texttt{New} & \texttt{All} & \texttt{Old} & \texttt{New} & \texttt{All} & \texttt{Old} & \texttt{New} & \texttt{All} & \texttt{Old} & \texttt{New} \\
\midrule
ViT \cite{vit}                  & 18.88                 & 20.86 & 15.79 & 30.34             & 35.42 & 21.83 & 29.52                 & 32.76 & 24.85 & 14.96             & 15.6  & 14.12 & 18.59                     & 20.12 & 16.4  & 30.39                 & 32.51 & 26.84 \\
GCD \cite{gcd}                  & 31.65                 & 32.11 & 30.93 & 63.18             & 64.35 & 61.22 & 63.85                 & 66.56 & 59.96 & 51.96             & 52.7  & 51    & 62.62                     & 65.29 & 58.79 & 60.59                 & 67.13 & 49.61 \\
SimGCD \cite{gcd3}               & 24.54                 & 34.35 & 8.09  & 41.95             & 57.92 & 13.54 & 46.78                 & 65.54 & 14.73 & 31.11             & 39.56 & 11.88 & 25.66                     & 37.66 & 5.15  & 28.88                 & 41.38 & 12.96 \\
CDAD-Net \cite{rongali2024cdadnetbridgingdomaingaps}             & 30.95 & 33.65 & 26.43           & 64.99 & 68.04 & 59.32                   & 67.5  & \cellcolor{red!18}70.89 & 61.72               & 53.36 & \cellcolor{yellow!20}56.05 & 47.23              & 64.7  & \cellcolor{yellow!20}69.4  & 55.25                  & \cellcolor{red!18}67.02 & \cellcolor{yellow!20}68.8  & 63.7                    \\
GCD With Synthetic~  & 29.86                 & 31.04 & 28.02 & 57.92             & 63.12 & 49.19 & 59.47                 & 59.59 & 59.29 & 53.3              & 52.84 & \cellcolor{yellow!20}53.89 & 61.46                     & 58.27 & 66.06 & 63.84                 & 64.04 & 63.51 \\
CDAD-Net with Synthetic  & \cellcolor{yellow!20}31.97 & \cellcolor{yellow!20}35.1  & 26.71           & \cellcolor{yellow!20}65.39 & \cellcolor{red!18}68.94 & 62.51                   & \cellcolor{red!18}67.83 & \cellcolor{yellow!20}70.87 & 62.64               & \cellcolor{red!18}53.51 & \cellcolor{red!18}56.65 & 46.37              & \cellcolor{red!18}66.97 & \cellcolor{red!18}69.76 & 62.2                   & 61.4  & 65.55 & 57.4  \\
{\ourmodel}(TIES-Merging\cite{yadav2024ties}) & \cellcolor{red!18}33.96                 & \cellcolor{red!18}37    & 29.23 & 59.99             & 62.93 & 55.07 & 66.26                 & 68.42 & 63.15 & 52.18             & 52.3  & 52.04 & 58.16                     & 58.62 & 57.5  & 65.32                 & \cellcolor{red!18}72.33 & 53.56 \\
{\ourmodel}[TA\cite{task_arithmetic}]           & 29.52                 & 27.31 & \cellcolor{red!18}33.06 & 62.42             & 61.67 & 63.59 & 64.46                 & 62.14 & 67.8  & 51.24             & 53.32 & 47.12 & 64.23                     & 61.24 & 68.92 & 65.28                 & 66.03 & \cellcolor{yellow!20}64.13 \\
{\ourmodel}(Ours)         & 31.51                 & 31.96 & 30.81 & \cellcolor{red!18}67.46             & \cellcolor{yellow!20}68.73 & \cellcolor{red!18}65.32 & 64.45                 & 60.25 & \cellcolor{yellow!20}70.48 & 50.76             & 48.76 & 53.36 & 64.77                     & 60.58 & \cellcolor{yellow!20}70.79 & 65.34                 & 67.48 & 61.76 \\
{\ourmodel}(Ours)[LoRA \cite{hu2021lora}]   & 31.56                 & 31.85 & \cellcolor{yellow!20}31.1  & 65.22             & 65.68 & \cellcolor{yellow!20}64.45 & \cellcolor{yellow!20}67.81                 & 65.14 & \cellcolor{red!18}71.66 & \cellcolor{yellow!20}53.4              & 48.47 & \cellcolor{red!18}59.81 & \cellcolor{yellow!20}66.13                     & 61.63 & \cellcolor{red!18}72.61 & \cellcolor{yellow!20}66.16                 & 67.12 & \cellcolor{red!18}64.57 \\
\bottomrule
\end{tabular}}

\vspace{1px}

\resizebox{0.98\textwidth}{!}{%
\begin{tabular}{>{\bfseries}lccc|ccc|ccc|ccc|ccc|ccc}
\midrule
 \multirow{2}{*}{\textbf{Methods}} & \multicolumn{3}{c|}{\textbf{Product $\rightarrow$ Art}} & \multicolumn{3}{c|}{\textbf{Product $\rightarrow$ Real World}} & \multicolumn{3}{c|}{\textbf{Product $\rightarrow$ Clipart}} & \multicolumn{3}{c|}{\textbf{Real World $\rightarrow$ Art}} & \multicolumn{3}{c|}{\textbf{Real World $\rightarrow$ Product}} & \multicolumn{3}{c}{\textbf{Real World $\rightarrow$ Clipart}} \\
 \cline{2-4} \cline{5-7} \cline{8-10} \cline{11-13} \cline{14-16} \cline{17-19}
 & \texttt{All} & \texttt{Old} & \texttt{New} & \texttt{All} & \texttt{Old} & \texttt{New} & \texttt{All} & \texttt{Old} & \texttt{New} & \texttt{All} & \texttt{Old} & \texttt{New} & \texttt{All} & \texttt{Old} & \texttt{New} & \texttt{All} & \texttt{Old} & \texttt{New} \\
\midrule
ViT \cite{vit}                  & 23.2                 & 24.64 & 21.33 & 31.21               & 35.45 & 25.13 & 19.27           & 20.52 & 17.31 & 32.22          & 35.79 & 27.58 & 44.67              & 52.21 & 32.03 & 20.8               & 23.71 & 16.26 \\
GCD \cite{gcd}                  & 50.27                & 48.18 & 52.99 & 65.07               & 63.09 & 67.91 & 29.08           & 29.22 & 28.87 & 54.26          & 54.05 & 54.55 & \cellcolor{red!18}69.04              & \cellcolor{yellow!20}72.76 & \cellcolor{yellow!20}62.79 & 31.04              & 34.93 & 24.97 \\
SimGCD \cite{gcd3}               & 38.28                & 50.42 & 10.66 & 48.36               & 67.07 & 16.41 & 22.45           & 32.37 & 11.34 & 48.95          & \cellcolor{red!18}66.79 & 8.36  & 57.19                   & 69.23       & 44.15       & 21.7               & 31.46 & 5.33  \\
CDAD-Net \cite{rongali2024cdadnetbridgingdomaingaps}                & 50.1  & 52.43 & 44.67           & 66.47 & \cellcolor{red!18}72.13 & 56.81                   & 31.36 & \cellcolor{yellow!20}34.6  & 25.94               & \cellcolor{yellow!20}54.68 & 58.07 & 46.96              & 61.39 & 64.79 & 55.06                  & 31.78 & \cellcolor{yellow!20}36.02 & 24.69                   \\

GCD With Synthetic~  & 49.18                & 46.54 & 52.61 & 63.4                & 59.67 & \cellcolor{yellow!20}68.77 & 28.43           & 27.72 & 29.55 & 51.71          & 61.55 & 38.91 & 61.14              & 65.34 & 54.1  & 26.38              & 28.11 & 23.68 \\
CDAD-Net with Synthetic & \cellcolor{red!18}54.12 & \cellcolor{red!18}57.67 & 46.04           & \cellcolor{yellow!20}66.97 & \cellcolor{yellow!20}70.2  & 61.46                   & \cellcolor{red!18}32.34 & \cellcolor{red!18}35.13 & 28.68               & 53.72 & 56.89 & 46.5               & 56.47 & 62.33 & 45.62                  & 31.19 & 33.67 & 27.02  \\
{\ourmodel}(TIES-Merging\cite{yadav2024ties}) & \cellcolor{yellow!20}53.28                & \cellcolor{yellow!20}54.77 & 51.33 & 62.74               & 66.85 & 56.83 & \cellcolor{yellow!20}31.82           & 33.97 & 28.46 & \cellcolor{red!18}57.11          & \cellcolor{yellow!20}66.14 & 45.36 & 67.04              & \cellcolor{red!18}74.25 & 54.95 & \cellcolor{red!18}34.41              & \cellcolor{red!18}37.94 & 28.9  \\
{\ourmodel}[TA\cite{task_arithmetic}]           & 49.92                & 52.33 & 45.17 & 65.57               & 67.22 & 62.99 & 31.48           & 30.21 & \cellcolor{red!18}33.51 & 51.65          & 55.06 & 44.92 & 65.01              & 63.73 & \cellcolor{red!18}66.99 & 30.73              & 28.65 & \cellcolor{yellow!20}34.08 \\
{\ourmodel}(Ours)         & 52.45                & 50.51 & \cellcolor{red!18}54.98 & \cellcolor{red!18}67.87               & 69.88 & 64.97 & 30.71           & 30.05 & \cellcolor{yellow!20}31.75 & 52.31          & 49.42 & \cellcolor{red!18}56.07 & \cellcolor{yellow!20}67.37              & 71.65 & 60.19 & 31.28              & 31.13 & 31.51 \\
{\ourmodel}* (Ours)[LoRA\cite{hu2021lora}]   & 52.66                & 51.75 & \cellcolor{yellow!20}53.84 & 65.48               & 62    & \cellcolor{red!18}70.48 & 31.52           & 31.83 & 31.04 & 53.42          & 51.6  & \cellcolor{yellow!20}55.78 & 66.33              & 68.97 & 61.91 & \cellcolor{yellow!20}32.26              & 30.4  & \cellcolor{red!18}35.15 \\
\bottomrule
\end{tabular}}
\caption{Detailed comparison of our proposed {\ourmodel} on DG-GCD with respect to referred literature for Office-Home Dataset}
\label{tab:office_home_comparison2}
\end{table}

\begin{table}[!htbp]
\centering
\resizebox{0.98\textwidth}{!}{%
\begin{tabular}{>{\bfseries}lccc|ccc|ccc|ccc|ccc}
\toprule
 \multicolumn{16}{c}{\textbf{DomainNet}} \\ 
\midrule
 \multirow{2}{*}{\textbf{Methods}} & \multicolumn{3}{c|}{\textbf{Sketch $\rightarrow$ Real}} & \multicolumn{3}{c|}{\textbf{Sketch $\rightarrow$ Quickdraw}} & \multicolumn{3}{c|}{\textbf{Sketch $\rightarrow$ Infograph}} & \multicolumn{3}{c|}{\textbf{Sketch $\rightarrow$ Painting}} & \multicolumn{3}{c}{\textbf{Sketch $\rightarrow$ Clipart}} \\
 \cline{2-4} \cline{5-7} \cline{8-10} \cline{11-13} \cline{14-16}
 & \texttt{All} & \texttt{Old} & \texttt{New} & \texttt{All} & \texttt{Old} & \texttt{New} & \texttt{All} & \texttt{Old} & \texttt{New} & \texttt{All} & \texttt{Old} & \texttt{New} & \texttt{All} & \texttt{Old} & \texttt{New} \\
\midrule
ViT \cite{vit}                  & 47.17              & 47.92 & 44.95 & 12.13              & 12.1  & 12.21 & 11.99              & 12.68 & 10.28 & 30.95             & 33.02 & 25.75 & \cellcolor{red!18}32.64             & \cellcolor{red!18}34.29 & \cellcolor{yellow!20}28.64 \\
GCD \cite{gcd}                  & 51.13              & 51.88 & 48.92 & \cellcolor{red!18}16.08              & \cellcolor{red!18}15.65 & \cellcolor{red!18}17.2  & 12.6               & 12.57 & \cellcolor{yellow!20}12.68 & 35.25             & 35.96 & \cellcolor{yellow!20}33.46 & \cellcolor{yellow!20}31.22             & 30.85 & \cellcolor{red!18}32.1  \\
SimGCD \cite{gcd3}               & 3.11               & 3.47  & 2.32  & 2.31               & 2.4   & 2.1   & 3.16               & 2.27  & 5.24  & 4.1               & 2.57  & 5.62  & 3.02              & 2.3   & 4.07  \\
CDAD-Net \cite{rongali2024cdadnetbridgingdomaingaps}                 & 48.21              & 47.7  & \cellcolor{yellow!20}49.77 & 12.27              & 11.52 & 14.24 & 12.07              & 12.69 & 11.34 & 35.47             & 36.39 & 32.86 & 18.63             & 17.52 & 20.39 \\
GCD With Synthetic~  & 53                 & 51.71 & 47.64 & 13.71              & \cellcolor{yellow!20}13.79 & 13.99 & 12.24              & 11.99 & 11.37 & 35.43             & 34.12 & 30.83 & 22.49             & 22.2  & 21.49 \\
CDAD-Net with Synthetic  & 47.11              & 46.09 & 49.4  & 12.75              & 13.1  & 14.05 & 12.52              & 13.04 & 11.92 & 35.87             & 36.73 & 33.35 & 18.99             & 17.68 & 21.07 \\
{\ourmodel} (TIES-Merging\cite{yadav2024ties}) & 50.32              & 52.88 & 42.8  & 15.22              & 15.12 & 15.49 & \cellcolor{red!18}14.75              & \cellcolor{red!18}16.04 & 11.53 & 35.84             & \cellcolor{yellow!20}38.99 & 27.93 & 31.06             & \cellcolor{yellow!20}33.34 & 25.53 \\
{\ourmodel} [TA\cite{task_arithmetic}]           & 51.84              & 52.58 & 49.65 & 13.67              & 13.44 & 14.25 & 12.72              & 13.05 & 11.89 & 33.96             & 35.32 & 30.55 & 21.94             & 21.8  & 22.29 \\
{\ourmodel} (Ours)         & \cellcolor{red!18}53.67              & \cellcolor{red!18}55.48 & 48.35 & 15.9               & 16    & \cellcolor{yellow!20}15.63 & \cellcolor{yellow!20}14.63              & \cellcolor{yellow!20}15.66 & 12.06 & \cellcolor{red!18}37.44             & \cellcolor{red!18}39.53 & 32.19 & 30.47             & 32.89 & 24.58 \\
{\ourmodel} * (Ours)[LoRA\cite{hu2021lora}]   & \cellcolor{yellow!20}53.01              & \cellcolor{yellow!20}53.75 & \cellcolor{red!18}50.84 & \cellcolor{yellow!20}13.71              & 13.38 & 14.57 & 13.82              & 14.23 & \cellcolor{red!18}12.78 & \cellcolor{yellow!20}36.77             & 37.9  & \cellcolor{red!18}33.93 & 24.17             & 24.46 & 23.44 \\

\bottomrule
\end{tabular}}

\resizebox{0.98\textwidth}{!}{%
\begin{tabular}{>{\bfseries}lccc|ccc|ccc|ccc|ccc}
\midrule
 \multirow{2}{*}{\textbf{Methods}} & \multicolumn{3}{c|}{\textbf{Clipart $\rightarrow$ Infograph}} & \multicolumn{3}{c|}{\textbf{Clipart $\rightarrow$ Quickdraw}} & \multicolumn{3}{c|}{\textbf{Clipart $\rightarrow$ Sketch}} & \multicolumn{3}{c|}{\textbf{Clipart $\rightarrow$ Real}} & \multicolumn{3}{c}{\textbf{Clipart $\rightarrow$ Painting}} \\
 \cline{2-4} \cline{5-7} \cline{8-10} \cline{11-13} \cline{14-16}
 & \texttt{All} & \texttt{Old} & \texttt{New} & \texttt{All} & \texttt{Old} & \texttt{New} & \texttt{All} & \texttt{Old} & \texttt{New} & \texttt{All} & \texttt{Old} & \texttt{New} & \texttt{All} & \texttt{Old} & \texttt{New} \\
\midrule
ViT \cite{vit}                  & 12.18                   & 12.64 & 11.03 & 12.13                   & 12.1  & 12.21 & 24.76                & 26.24 & \cellcolor{red!18}21.27 & 44.14              & 45.43 & 40.34 & 26.76                  & 28.7  & 21.91 \\
GCD \cite{gcd}                  & 14.03                   & 14.64 & 12.49 & \cellcolor{red!18}14.94                   & \cellcolor{yellow!20}14.67 & \cellcolor{yellow!20}15.65 & 25.33                & 27.68 & 19.78 & 53.23              & 55.48 & 46.62 & 34.82                  & 36.82 & 29.83 \\
SimGCD \cite{gcd3}               & 2.03                    & 0.4   & 3.94  & 0.5                     & 0.3   & 1     & 1                    & 0.02  & 3.842 & 1.64               & 1.07  & 2.42  & 2.07                   & 2.05  & 2.13  \\
CDAD-Net \cite{rongali2024cdadnetbridgingdomaingaps}                 & 12.79                   & 12.96 & \cellcolor{red!18}12.87 & 12.06                   & 11.59 & 12.78 & 19                   & 19.17 & 18.76 & 47.06              & 44.62 & 49.2  & 34.45                  & 36.02 & 32.85 \\
GCD With Synthetic~  & 11.46                   & 12.03 & 10.04 & 12.68                   & 12.57 & 12.95 & 18.74                & 20.54 & 14.47 & 50.11              & 52.26 & 43.79 & 32.67                  & 34.91 & 27.06 \\
CDAD-Net with Synthetic  & 13                      & 13.37 & 12.56 & 12.07                   & 11.76 & 12.89 & 17.46                & 18.03 & 16.67 & 48.25              & 47.51 & 49.6  & 33.23                  & 32.79 & \cellcolor{yellow!20}34.2  \\
{\ourmodel} (TIES-Merging\cite{yadav2024ties}) & 15.66                   & \cellcolor{yellow!20}17.02 & 12.28 & \cellcolor{yellow!20}14.91                   & \cellcolor{red!18}14.73 & 15.39 & \cellcolor{red!18}27.75                & \cellcolor{red!18}30.64 & \cellcolor{yellow!20}20.89 & \cellcolor{yellow!20}54.18              & \cellcolor{red!18}56.73 & 46.72 & 36.71                  & 38.14 & 33.16 \\
{\ourmodel} [TA\cite{task_arithmetic}]           & \cellcolor{yellow!20}15.71                   & 16.88 & \cellcolor{yellow!20}12.78 & 14.63                   & 14.18 & \cellcolor{red!18}15.81 & \cellcolor{yellow!20}27.03                & \cellcolor{yellow!20}29.89 & 20.26 & 53.91              & 55.14 & \cellcolor{yellow!20}50.29 & \cellcolor{yellow!20}36.85                  & \cellcolor{red!18}39.69 & 29.75 \\
{\ourmodel} (Ours)         & \cellcolor{red!18}15.81                   & \cellcolor{red!18}17.09 & 12.63 & 14.53                   & 14.14 & 15.58 & 26.86                & 29.49 & 20.64 & \cellcolor{red!18}54.54              & \cellcolor{yellow!20}56.03 & 50.17 & 36.81                  & \cellcolor{yellow!20}38.87 & 31.67 \\
{\ourmodel}* (Ours)[LoRA\cite{hu2021lora}]   & 14.19                   & 14.12 & 14.35 & 13.31                   & 13.23 & 13.53 & 22.01                & 22.9  & 19.91 & 53.95              & 54.99 & \cellcolor{red!18}50.91 & \cellcolor{red!18}37.12                  & 37.89 & \cellcolor{red!18}35.21 \\
\bottomrule
\end{tabular}}

\vspace{1px}
\resizebox{0.98\textwidth}{!}{%
\begin{tabular}{>{\bfseries}lccc|ccc|ccc|ccc|ccc}
\midrule
 \multirow{2}{*}{\textbf{Methods}} & \multicolumn{3}{c|}{\textbf{Painting $\rightarrow$ Infograph}} & \multicolumn{3}{c|}{\textbf{Painting $\rightarrow$ Quickdraw}} & \multicolumn{3}{c|}{\textbf{Painting $\rightarrow$ Sketch}} & \multicolumn{3}{c|}{\textbf{Painting $\rightarrow$ Real}} & \multicolumn{3}{c}{\textbf{Painting $\rightarrow$ Clipart}} \\
 \cline{2-4} \cline{5-7} \cline{8-10} \cline{11-13} \cline{14-16}
 & \texttt{All} & \texttt{Old} & \texttt{New} & \texttt{All} & \texttt{Old} & \texttt{New} & \texttt{All} & \texttt{Old} & \texttt{New} & \texttt{All} & \texttt{Old} & \texttt{New} & \texttt{All} & \texttt{Old} & \texttt{New} \\
\midrule
ViT \cite{vit}                  & 12.2                     & 13.1  & 9.94  & 12.13                    & 12.1  & 12.21 & 23                    & 24.78 & 18.79 & 51.53               & 54.16 & 43.8  & 26.57                  & 28.08 & 22.92 \\
GCD \cite{gcd}                  & 12.87                    & 12.67 & \cellcolor{yellow!20}13.37 & 10.74                    & 10.56 & 11.21 & 21.49                 & 22.26 & \cellcolor{red!18}19.68 & 52.12               & 51.86 & \cellcolor{red!18}52.86 & 25.32                  & 24.79 & \cellcolor{red!18}26.6  \\
SimGCD \cite{gcd3}               & 3.2                      & 2.6   & 3.8   & 3.5                      & 2.32  & 4.65  & 4.23                  & 3.56  & 4.86  & 4.2                 & 3.52  & 5     & 4.49                   & 3.6   & 5.23  \\
CDAD-Net \cite{rongali2024cdadnetbridgingdomaingaps}                 & 11.65                    & 12.49 & 10.66 & 11.98                    & 11.2  & 12.44 & 17.11                 & 17.68 & 16.32 & 49.04               & 48.63 & 50.27 & 20.06                  & 19.74 & 20.57 \\
GCD With Synthetic  & 10.86                    & 10.56 & 9.84  & 11.81                    & 11.8  & 11.77 & 17.26                 & 16.25 & 13.83 & 49.1                & 47.3  & 42.04 & 19.3                   & 19.45 & 18.04 \\
CDAD-Net with Synthetic  & 11.53                    & 12.32 & 10.59 & 11.86                    & 10.71 & 12.32 & 17.29                 & 18.45 & 15.7  & 48.4                & 50.23 & 49.7  & 17.44                  & 15.92 & 19.86 \\
{\ourmodel} (TIES-Merging\cite{yadav2024ties}) & \cellcolor{yellow!20}15.34                    & \cellcolor{yellow!20}16.64 & 12.13 & 12.89                    & 12.64 & \cellcolor{red!18}13.58 & \cellcolor{red!18}23.45                 & \cellcolor{yellow!20}25.6  & 18.38 & \cellcolor{yellow!20}55.16               & \cellcolor{yellow!20}57.3  & 47.46 & \cellcolor{yellow!20}27.5                   & \cellcolor{red!18}29.48 & 22.65 \\
{\ourmodel} [TA\cite{task_arithmetic}]           & 15.17                    & 16.52 & 11.8  & 12.78                    & 12.58 & 13.29 & \cellcolor{yellow!20}23.21                 & \cellcolor{red!18}25.69 & 17.34 & \cellcolor{red!18}55.16               & \cellcolor{red!18}57.31 & 48.87 & 26.76                  & 27.91 & 23.96 \\
{\ourmodel} (Ours)         & \cellcolor{red!18}15.71                    & \cellcolor{red!18}16.72 & 13.22 & \cellcolor{yellow!20}12.9                     & \cellcolor{yellow!20}12.66 & \cellcolor{yellow!20}13.53 & 23.14                 & 25.23 & 18.19 & 55.07               & 56.97 & 49.5  & \cellcolor{red!18}27.6                   & \cellcolor{yellow!20}29.07 & 24.03 \\
{\ourmodel}* [Ours)[LoRA\cite{hu2021lora}]   & 14.41                    & 14.68 & \cellcolor{red!18}13.74 & \cellcolor{red!18}12.9                     & \cellcolor{red!18}12.9  & 12.91 & 21.39                 & 22.39 & \cellcolor{yellow!20}19.03 & 53.83               & 54.99 & \cellcolor{yellow!20}50.44 & 22.94                  & 22.41 & \cellcolor{yellow!20}24.24 \\
\bottomrule
\end{tabular}}
\caption{Detailed comparison of our proposed {\ourmodel} on DG-GCD with respect to referred literature for DomainNet Dataset}
\label{tab:domain_net_comparison3}
\end{table}

\newpage

\section{Performance comparison with Domain Adaptation (DA) methods} \label{sec:performance_comparison} Table \ref{tab:results_crow} presents a performance comparison of our proposed method, {{\ourmodel}}, against two prominent methods, CROW (DA) and CDAD-Net (DA), across three benchmark datasets: PACS, Office-Home, and DomainNet. Notably, CDAD-Net (DA) represents a strong upper bound as it operates in the Domain Adaptation (DA) setting, leveraging access to target-domain data during training. In contrast, {{\ourmodel}} is designed for the more challenging Domain Generalization (DG) setting, where no target-domain information is available. While CROW (DA) is included in this supplementary comparison for reference, CDAD-Net (DA) serves as a more appropriate upper bound, given its superior performance.

Our method significantly outperforms CROW (DA) across all datasets in terms of overall accuracy, achieving a margin of +9.34\% on PACS (73.30\% vs. 63.96\%), +3.39\% on Office-Home (53.86\% vs. 50.47\%), and +2.81\% on DomainNet (29.01\% vs. 26.20\%). Additionally, {{\ourmodel}} demonstrates robust generalization across both old and new classes, underscoring its adaptability in diverse scenarios. While CDAD-Net (DA) achieves higher performance due to its reliance on target domain data, the comparison highlights the inherent trade-off between the DA and DG settings. By including CROW (DA) results here, we provide a holistic view of baseline performance while emphasizing the relevance of CDAD-Net (DA) as the key upper bound in this context. This reinforces the practical value of {{\ourmodel}} in solving the domain generalization challenge without relying on target domain assumptions.

\begin{table}[!htbp]
  \centering
  \resizebox{0.9\textwidth}{!}{%
  \begin{tabular}{llcccc|ccc|ccccc}
    \toprule
    \multirow{2}{*}{\textbf{Methods}} & \multirow{2}{*}{\textbf{Venue}} & \multirow{2}{*}{\textbf{Target-Domain}} 
    & \multicolumn{3}{c|}{\textbf{PACS}} & \multicolumn{3}{c|}{\textbf{Office-Home}} & \multicolumn{3}{c}{\textbf{DomainNet}} \\
    \cmidrule(lr){4-6}
    \cmidrule(lr){7-9}
    \cmidrule(lr){10-12}
    & & & \textbf{All} & \textbf{Old} & \textbf{New} & \textbf{All} & \textbf{Old} & \textbf{New} & \textbf{All} & \textbf{Old} & \textbf{New} \\
    \midrule
    \textbf{CROW (DA) \cite{wen2024cross}} & ICML'24 & $\checkmark$ & 63.96 & 61.78 & 68.65 & 50.47 & 54.50 & 39.71 & 26.20 & 26.60 & 25.80 \\
    \textbf{{{\ourmodel}} (Ours)} & -- & $\times$ & 73.30 & 75.28 & 72.56 & 53.86 & 53.37 & 54.33 & 29.01 & 30.38 & 25.46  \\
    \hline 
    \textbf{CDAD-Net (DA) \cite{rongali2024cdadnetbridgingdomaingaps} [Upper bound]} & CVPR-W'24  & $\checkmark$ & 83.25 & 87.58 & 77.35 & 67.55 & 72.42 & 63.44 & 70.28 & 76.46 & 65.19  \\
    \bottomrule
  \end{tabular}}
  \caption{\textbf{Performance comparison} of CROW method with our method, as well as the upper bound CDAD-Net (DA), on all datasets.}
  \label{tab:results_crow}
\end{table}

\section{Additional Ablation Studies} \label{sec:ablations}

\noindent \textbf{Component Impact on Office-Home}: Table \ref{tab:office_home_ablation} reveals the significant effects of essential components on the DG$^2$CD-Net's overall performance. Removing synthetic domains leads to a decrease of approximately \(3.28\%\) in the ``\texttt{All}'' metric, underscoring its importance for generalization. The absence of episodic training results in a decrease of \(2.33\%\), highlighting its role in model adaptability. The most considerable impact is observed with a fixed old/novel class split, which shows a reduction of \(4.67\%\) compared to the full model's configuration. In contrast, the full implementation of DG$^2$CD-Net achieves a comprehensive performance of \(53.79\%\) across all classes, demonstrating the effectiveness of dynamic weighting and the combined utility of all components in enhancing domain generalization and class discovery in the Office-Home dataset.

\begin{table}[!htbp]
    \centering
    \begin{tabular}{lccc}
        \toprule
        \multirow{2}{*}{\textbf{Model Variant}} & \multicolumn{3}{c}{\textbf{Office-Home}} \\
        \cline{2-4}
        & \textbf{All} & \textbf{Old} & \textbf{New} \\
        \midrule
        \checkmark \textbf{Without Synthetic Domain} & 50.51 & 50.58 & 50.31 \\
        \checkmark \ \textbf{Without multi-global updates} & 51.46 & 50.77 & 52.26  \\
        \checkmark \ \textbf{Static known/novel class split across episodes} & 49.11 & \cellcolor{red!18}56.23 & 38.85 \\
        \midrule
        \rowcolor{blue!10}
        \textbf{Full DG$^2$CD-Net (Proposed)} & \textbf{\cellcolor{red!18}53.79} & \textbf{53.83} & \textbf{\cellcolor{red!18}53.66} \\
        \bottomrule
    \end{tabular}
    \caption{Ablation study results on the impact of various components of DG$^2$CD-Net for the Office-Home dataset.}
    \label{tab:office_home_ablation}
\end{table}

\noindent \textbf{Observation on Old-New class splits:}
Table \ref{tab:splits} illustrates the impact of varying base (old) and novel (new) class splits on the performance of {\ourmodel} on the PACS dataset. We tested five different splits, ranging from 2 old classes with five novel classes to 6 old classes with one novel class. The results show that as the proportion of novel classes increases, the model's performance on novel classes improves, but there is a slight decline in accuracy for base classes. This behavior highlights the challenge of maintaining a balance between recognizing old and novel categories. The best overall performance is observed with the 5-2 split, indicating that DG$^{2}$CD-Net is more effective when the distribution of old and novel classes is moderately balanced.

\begin{table}[!htbp]
    \centering
    \begin{tabular}{cccc}
        \toprule
        \multirow{2}{*}{\textbf{Splits (Old-New)}} & \multicolumn{3}{c}{\textbf{PACS}} \\
        \cline{2-4}
        & \textbf{All} & \textbf{Old} & \textbf{New} \\
        \midrule
        \textbf{2 - 5} & 74.94 & 73.29 & 76.77 \\
        \textbf{3 - 4} & 75.11 & 76.28 & 74.69 \\
        \textbf{4 - 3} & 73.3 & 75.28 & 72.56 \\
        \textbf{5 - 2} & 75.85 & 74.01 & 80.72 \\
        \textbf{6 - 1} & 74.89 & 74.90 & 74.96 \\
        \bottomrule
    \end{tabular}
    \caption{Sensitivity on different \texttt{Old}-\texttt{New} class splits.}
    \label{tab:splits}
\end{table}

\begin{figure}[!htbp]
\centering
    \begin{tikzpicture}
        \begin{axis}[
            ybar,
            bar width=10pt,
            symbolic x coords={All, Old, New}, 
            xtick=data,
            ymin=65, ymax=80,
            enlarge x limits=0.2,
            legend style={at={(0.5,0.97)}, anchor=north ,legend columns=-1, font=\footnotesize, draw=black, /tikz/every even column/.append style={column sep=5pt}},
            ylabel={Accuracy},
            ylabel style={font=\small},
            xlabel style={font=\small},
            nodes near coords,
            every node near coord/.append style={font=\tiny}, 
            width=8.5cm, 
            height=5cm, 
            legend image code/.code={
                \draw[#1,fill=#1,draw=black] (0cm,-0.1cm) rectangle (0.2cm,0.2cm);
            }
        ]
        \addplot+[ybar, fill=red, draw=none, nodes near coords, every node near coord/.append style={color=red}] 
            coordinates {(All,71.45) (Old,72.92) (New,70.33)};
        
        \addplot+[ybar, fill=blue, draw=none, nodes near coords, every node near coord/.append style={color=blue}] 
            coordinates {(All,73.3) (Old,75.28) (New,72.56)};
        
        \legend{Without Marginal loss, With Marginal loss}    
        \end{axis}
    \end{tikzpicture}
    \caption{Accuracy comparison of Old, All, and New categories with and without $\mathcal{L}_{\text{margin}}$ loss.}
    \label{fig:marginal_loss}
\end{figure}
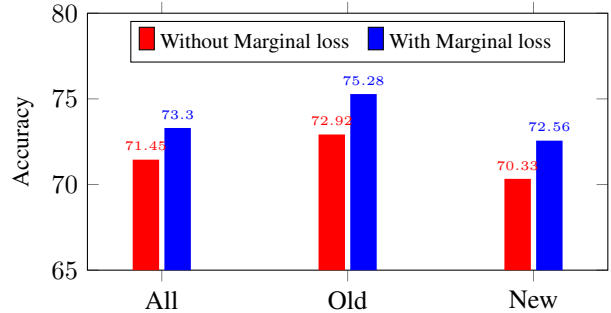

\vspace{10pt}
\noindent \textbf{Effect of $\boldsymbol{\mathcal{L}_{\textbf{margin}}}$ :}
As shown in Figure~\ref{fig:marginal_loss}, adding \(\mathcal{L}_{\text{margin}}\) improves accuracy across all categories. \texttt{Old} classes benefit the most with a 2.36\% increase, indicating enhanced feature separation for well-known classes. \texttt{New} classes see a 2.23\% improvement, suggesting better discrimination for novel classes. Overall, accuracy increases by 1.85\%, demonstrating that \(\mathcal{L}_{\text{margin}}\) enhances class separability and generalization across both familiar and unseen categories.

\noindent \textbf{Effect of} $\boldsymbol{m}$ \textbf{in} $\boldsymbol{\mathcal{L}_{\textbf{margin}}:}$ Table \ref{tab:margin} illustrates the average accuracy of our model across different margin values (m) in $\mathcal{L}_{margin}$. We experimented with different values of m ranging from 0.3 to 0.9. The highest scores were observed for $m$ = 0.7, indicating that this margin setting provides better separation between known and novel classes.

\begin{figure}[!ht]
    \centering
    \begin{minipage}[c]{0.45\textwidth}
        \centering
        \begin{tabular}{cccc}
            \toprule
            \multirow{2}{*}{\textbf{\(\mathbf{m}\)}} & \multicolumn{3}{c}{\textbf{PACS}} \\
            \cline{2-4}
            & \textbf{All} & \textbf{Old} & \textbf{New} \\
            \midrule
            \textbf{0.3} & 71.50 & 74.48 & 69.05 \\
            \textbf{0.4} & 71.27 & 75.58 & 67.92 \\
            \textbf{0.5} & 69.82 & 74.00 & 66.23 \\
            \textbf{0.6} & 63.13 & 65.42 & 61.99 \\
            \textbf{0.7} & \cellcolor{red!18}73.30 & 75.28 & \cellcolor{red!18}72.56 \\
            \textbf{0.8} & 72.24 & \cellcolor{red!18}75.77 & 69.92 \\
            \textbf{0.9} & 71.91 & 75.69 & 68.98 \\
            \bottomrule
        \end{tabular}
        \captionof{table}{Average accuracy for different sensitivity of the hyper-parameter \(m\).}
        \label{tab:margin}
    \end{minipage}%
    \hspace{0.05\textwidth} 
    \begin{minipage}[c]{0.45\textwidth}
        \centering
        \resizebox{\textwidth}{!}{
        \begin{tikzpicture}
            \begin{axis}[
                width=9cm, 
                height=5cm,
                xlabel={Value of \(m\)},
                xlabel style={font=\footnotesize},
                ylabel={Accuracy (\%)},
                ylabel style={font=\footnotesize},
                xmin=0.3, xmax=0.9,
                ymin=60, ymax=80,
                xtick={0.3, 0.4, 0.5, 0.6, 0.7, 0.8, 0.9},
                ytick={60, 65, 70, 75, 80},
                legend style={
                    font=\scriptsize,
                    draw=black,
                    fill=white,
                    at={(1.05,1)}, 
                    anchor=north west
                },
                grid=both,
                grid style={dashed, gray!30}
            ]
            \addplot[
                color=cvprblue,
                mark=*,
                ]
                coordinates {
                (0.3, 71.50) (0.4, 71.27) (0.5, 69.82) (0.6, 63.13) (0.7, 73.30) (0.8, 72.24) (0.9, 71.91)
                };
            \addlegendentry{All}
            
            \addplot[
                color=cvprgreen,
                mark=square*,
                ]
                coordinates {
                (0.3, 74.48) (0.4, 75.58) (0.5, 74.00) (0.6, 65.42) (0.7, 75.28) (0.8, 75.77) (0.9, 75.69)
                };
            \addlegendentry{Old}
            
            \addplot[
                color=red,
                mark=diamond*,
                ]
                coordinates {
                (0.3, 69.05) (0.4, 67.92) (0.5, 66.23) (0.6, 61.99) (0.7, 72.56) (0.8, 69.92) (0.9, 68.98)
                };
            \addlegendentry{New}
            
            \end{axis}
        \end{tikzpicture}
        }
        \caption{The relationship between margin \(m\) and accuracy (All, Old, New). The margin \(m\) in the loss function influences class separation, with peak accuracy observed at \(m = 0.7\).}
        \label{fig:margin_episode}
    \end{minipage}
\end{figure}

\section{Effect of initialization of backbone}  \label{sec:backbone_effect}

Backbone selection significantly impacts the performance of {\ourmodel} in domain generalization. Initially, we employed ViT-B/16 with DINO initialization for comparability with prior work, but inspired by DINO-v2’s advancements, we expanded our study to include ViT-B/16 with DINO-v2 initialization. Additionally, we have experimented with ResNet-50 with CLIP and ImageNet initializations and ViT-B/16 with CLIP initialization. Table \ref{tab:backbone} summarizes the results on PACS.

\begin{table}[!htbp]
    \centering
    \resizebox{0.4\linewidth}{!}{
    \begin{tabular}{ccccc}
    \toprule
    \multirow{2}{*}{\textbf{Model}} & \multirow{2}{*}{\textbf{Backbone}} & \multicolumn{3}{c}{\textbf{PACS}} \\
    \cline{3-5}
    & & \textbf{All} & \textbf{Old} & \textbf{New} \\
    \midrule
    \textbf{ResNet-50} &  \textbf{CLIP} \cite{radford2021learning} & 25.39 & 20.97 & 29.33 \\
    \textbf{ResNet-50} & \textbf{ImageNet} \cite{he2016deep} & 54.98 & 64.18 & 45.33 \\
    \textbf{ViT-B/16} & \textbf{DINO} \cite{dino} & 73.30 & 75.28 & 72.56 \\
    \textbf{ViT-B/16} & \textbf{DINO-v2} \cite{oquab2024dinov} & 87.71 & 90.67 & 
    84.91 \\
    \textbf{ViT-B/16} & \textbf{CLIP} \cite{radford2021learning} & \cellcolor{red!18}90.07 & \cellcolor{red!18}92.25 & \cellcolor{red!18}87.72 \\
    \bottomrule
    \end{tabular}}
    \caption{Performance Comparison of {\ourmodel} with different backbones on the PACS Dataset}
    \label{tab:backbone}
\end{table}

The results highlight the superiority of the DINO-v2 and CLIP-based models, with CLIP achieving the highest performance. The strong results of ViT-B/16 (CLIP) suggest that pre-training with vision-language data improves generalization. Given these findings, we recommend DINO-v2 and CLIP-based ViT as strong baselines for future domain generalization studies.

\section{Effect of LoRA Fine-Tuning}
\label{sec:LORA_effect}

In our experiments, we incorporated LoRA (Low-Rank Adaptation) into DG$^2$ CD-Net to improve the efficiency of fine-tuning while minimizing memory overhead. Unlike full fine-tuning, LoRA updates a small subset of parameters while keeping pre-trained weights frozen, thereby reducing catastrophic forgetting and enhancing adaptation and generalization. 

\begin{table}[h]
    \centering
    \begin{tabular}{l|c|c|c}
        \hline
        \textbf{Method} & \textbf{Trainable Parameters (K)} & \textbf{Total Parameters (K)} & \textbf{Percentage (\%)} \\
        \hline
        DG$^2$CD-Net (Vanilla) & 7,088 & 85,799 & 8.261 \\
        DG$^2$CD-Net* (LoRA) & 98 & 85,799 & 0.115 \\
        \hline
    \end{tabular}
    \caption{Comparison of model parameters \textit{with} and \textit{without} LoRA fine-tuning.}
    \label{tab:lora_comparison}
\end{table}

\vspace{2cm}
Table \ref{tab:dgcd_variants} presents a performance comparison of DG\textsuperscript{2}CD-Net using different LoRA-based adapters,  including \textit{LoRA} \cite{hu2021lora}, \textit{DoRA} \cite{pmlr-v235-liu24bn}, and \textit{AdaLoRA} \cite{zhang2023adaptive}. These adapters aim to improve the model efficiency while maintaining high accuracy, with LoRA achieving the best performance. \\

\begin{table}[!htbp]
    \centering
    \begin{tabular}{lc}
        \toprule
        \textbf{Adapters} & \textbf{All (\%)} \\
        \midrule
        \cellcolor{red!18}LoRA \cite{hu2021lora} & \cellcolor{red!18}75.21 \\
        DoRA \cite{pmlr-v235-liu24bn} & 74.11 \\
        AdaLoRA \cite{zhang2023adaptive} & 74.20 \\
        \bottomrule
    \end{tabular}
    \caption{Performance comparison of DG$^2$CD-Net with different LoRA Adapters}
    \label{tab:dgcd_variants}
\end{table}

These findings emphasize the efficiency of LoRA in reducing the computational burden of fine-tuning while maintaining the model’s overall capacity. Given the benefits, we recommend adopting LoRA-based fine-tuning in future research for domain generalization tasks to optimize memory usage and training speed without sacrificing performance.

\section{Limitations and Future Work} \label{sec:limitations} While {\ourmodel} demonstrates strong performance in domain generalization and category discovery, there are areas for future enhancement. One aspect that can be improved is the reliance on synthetic domain generation, which, while effective, can be optimized to reduce computational costs. Exploring more streamlined approaches to synthetic domain creation or alternative techniques that do not require synthetic data could further improve scalability and efficiency. Additionally, the episodic training framework, though beneficial for adaptation, demands considerable computational resources, especially when applied to large-scale datasets like DomainNet. Optimizing this process could make the method more feasible for real-world, large-scale applications.

In future work, efforts can focus on enhancing the efficiency of both synthetic domain generation and the episodic training process. Advanced techniques for model merging can also be explored to further improve performance. Addressing challenges like data imbalance, which is common in real-world scenarios, will strengthen the model's robustness and adaptability. Overall, extending {\ourmodel} in these directions holds great promise for developing even more scalable and effective solutions for complex tasks in domain generalization.

\newpage

{
    \small
    \bibliographystyle{ieeenat_fullname}
    \bibliography{main}
}

\end{document}